\documentclass{yresearch}
\usepackage{amsmath}
\usepackage{enumerate} 
\usepackage{algorithm}
\usepackage{algpseudocode}
\usepackage{amsfonts}
\usepackage{amsthm}
\usepackage{newtxtt}
\usepackage{cleveref}
\usepackage{diagbox} 
\usepackage{colortbl}
\usepackage{amssymb}
\usepackage{xspace}
\usepackage{wrapfig}
\usepackage{adjustbox}
\usepackage{needspace} 
\usepackage{tabularx}
\usepackage{booktabs}
\usepackage{mathtools}
\usepackage{tikz}
\usepackage{enumitem}
\usepackage{silence}
\usepackage{dsfont}
\usepackage[table]{xcolor}
\usepackage[dvipsnames]{xcolor}
\usepackage{multirow}
\usepackage{makecell}
\usepackage{xfakebold}

\usepackage{amsmath,amsfonts,bm}

\def\eqref#1{equation~\ref{#1}}

\def\1{\bm{1}}

\DeclareMathAlphabet{\mathsfit}{\encodingdefault}{\sfdefault}{m}{sl}
\SetMathAlphabet{\mathsfit}{bold}{\encodingdefault}{\sfdefault}{bx}{n}

\usepackage{multirow}
\usepackage[normalem]{ulem}
\usepackage{pifont}
\usepackage{cuted}

\useunder{\uline}{\ul}{}
\definecolor{AppleWhite}{RGB}{255,255,255}
\definecolor{ApplePrimaryCoolGray}{RGB}{116,128,139}
\definecolor{AppleCoolGray1}{RGB}{199,209,214}
\definecolor{AppleCoolGray2}{RGB}{147,174,190}
\definecolor{AppleCoolGray3}{RGB}{124,147,160}
\definecolor{AppleCoolGray4}{RGB}{92,102,109}
\definecolor{AppleCoolGray5}{RGB}{78,93,100}
\definecolor{AppleCoolGray6}{RGB}{53,60,65}
\definecolor{AppleBlack}{RGB}{0,0,0}
\definecolor{AppleSecondaryChartGray}{RGB}{168,168,168}
\definecolor{AppleChartGray2}{RGB}{233,233,233}
\definecolor{AppleChartGray3}{RGB}{211,211,211}
\definecolor{AppleChartGray4}{RGB}{190,190,190}
\definecolor{AppleChartGray5}{RGB}{140,140,140}
\definecolor{AppleChartGray6}{RGB}{102,102,102}
\definecolor{AppleChartGray7}{RGB}{64,64,64}
\definecolor{ApplePrimaryChartBlue}{RGB}{84,151,193}
\definecolor{AppleBlue2}{RGB}{212,229,239}
\definecolor{AppleBlue3}{RGB}{169,202,223}
\definecolor{AppleBlue4}{RGB}{127,177,209}
\definecolor{AppleBlue5}{RGB}{71,130,166}
\definecolor{AppleBlue6}{RGB}{55,99,128}
\definecolor{AppleBlue7}{RGB}{45,72,89}
\definecolor{ApplePrimaryChartGreen}{RGB}{83,172,121}
\definecolor{AppleGreen2}{RGB}{212,234,221}
\definecolor{AppleGreen3}{RGB}{169,213,188}
\definecolor{AppleGreen4}{RGB}{126,193,155}
\definecolor{AppleGreen5}{RGB}{58,140,82}
\definecolor{AppleGreen6}{RGB}{39,102,54}
\definecolor{AppleGreen7}{RGB}{29,58,31}
\definecolor{ApplePrimaryChartYellow}{RGB}{253,195,93}
\definecolor{AppleYellow2}{RGB}{254,240,214}
\definecolor{AppleYellow3}{RGB}{254,224,174}
\definecolor{AppleYellow4}{RGB}{254,210,134}
\definecolor{AppleYellow5}{RGB}{230,168,69}
\definecolor{AppleYellow6}{RGB}{191,131,46}
\definecolor{AppleYellow7}{RGB}{153,107,54}
\definecolor{ApplePrimaryChartOrange}{RGB}{250,151,92}
\definecolor{AppleOrange2}{RGB}{254,229,214}
\definecolor{AppleOrange3}{RGB}{252,203,173}
\definecolor{AppleOrange4}{RGB}{252,178,133}
\definecolor{AppleOrange5}{RGB}{227,121,68}
\definecolor{AppleOrange6}{RGB}{191,87,46}
\definecolor{AppleOrange7}{RGB}{143,59,36}
\definecolor{ApplePrimaryChartRed}{RGB}{227,94,105}
\definecolor{AppleRed2}{RGB}{248,215,217}
\definecolor{AppleRed3}{RGB}{241,174,180}
\definecolor{AppleRed4}{RGB}{234,135,143}
\definecolor{AppleRed5}{RGB}{196,63,77}
\definecolor{AppleRed6}{RGB}{153,35,53}
\definecolor{AppleRed7}{RGB}{102,19,43}
\definecolor{ApplePrimaryChartPurple}{RGB}{161,150,204}
\definecolor{ApplePurple2}{RGB}{231,228,242}
\definecolor{ApplePurple3}{RGB}{208,202,229}
\definecolor{ApplePurple4}{RGB}{185,176,217}
\definecolor{ApplePurple5}{RGB}{128,113,171}
\definecolor{ApplePurple6}{RGB}{89,76,128}
\definecolor{ApplePurple7}{RGB}{62,46,101}
\definecolor{AppleCoolGray}{RGB}{116,128,139}
\definecolor{AppleChartGray}{RGB}{168,168,168}
\definecolor{AppleBlue}{RGB}{84,151,193}
\definecolor{AppleGreen}{RGB}{83,172,121}
\definecolor{AppleYellow}{RGB}{253,195,93}
\definecolor{AppleOrange}{RGB}{250,151,92}
\definecolor{AppleRed}{RGB}{227,94,105}
\definecolor{ApplePurple}{RGB}{161,150,204}
\usepackage[table]{xcolor}
\usepackage{wrapfig}

\newcommand{\myparagraph}[1]{\vspace{1pt}\noindent{\bf{#1}}~~}
\newcommand{\titlefont}{\fontsize{19}{20}\selectfont}

\definecolor{myblue}{RGB}{0, 112, 192}
\definecolor{myred}{RGB}{192, 0, 0}

\definecolor{textgray}{HTML}{6E6E73}
\usetikzlibrary{positioning, calc}
\usetikzlibrary{decorations.pathmorphing}

\definecolor{hestbg}{HTML}{EEF4FC}      
\definecolor{hestavgbg}{HTML}{DCE9F8}

\definecolor{stimagebg}{HTML}{EEF7EE}   
\definecolor{stimageavgbg}{HTML}{DCEEDC}
\definecolor{stdgray}{HTML}{6F6F6F}
\definecolor{cartablue}{HTML}{1F5FBF}

\newcommand{\std}[1]{_{\textcolor{stdgray}{#1}}}
\newcommand{\res}[2]{#1\std{#2}}
\newcommand{\best}[2]{\mathbf{#1}\std{#2}}
\newcommand{\second}[2]{\underline{#1}\std{#2}}
\definecolor{finalblue}{RGB}{232,241,252}  
\definecolor{avgcol}{RGB}{246,248,250}     
\definecolor{linkblue}{RGB}{20,105,190}    
\definecolor{datasetheadbg}{HTML}{FFF3D6}   
\definecolor{datasetlabelbg}{HTML}{FFF8E8}  
\definecolor{datasettext}{HTML}{4F4A3F}
\definecolor{datasetrowbg}{HTML}{FFF1DF}  

\usepackage{wrapfig}
\makeatletter
\patchcmd{\wrong@fontshape}{\@gobbletwo}{}{}{}
\makeatother
\tcbuselibrary{minted}
\setminted[python]{
  linenos,
  breaklines,
  fontsize=\footnotesize,
  xleftmargin=2em
}

\makeatletter
\AtBeginDocument{
  \urlstyle{sf}
  
}
\makeatother

\definecolor{light}{RGB}{125, 125, 125}
\crefname{tcb@cnt@pbox}{code}{code}
\Crefname{tcb@cnt@pbox}{Code}{Code}
\crefname{assumption}{assumption}{assumption}
\Crefname{assumption}{Assumption}{Assumptions}

\newtcolorbox[auto counter]{pbox}[2][]{
  colback=white,
  title=Code~\thetcbcounter: #2,
  #1,fonttitle=\sffamily,
  fontupper=\sffamily,
  arc=2pt,
  colframe=bgcolor,
  coltitle=fgcolor,
  colbacktitle=bgcolor,
  toptitle=0.25cm,
  bottomtitle=0.125cm
}

\makeatletter
\newcommand\applefootnote[1]{%
  \begingroup
  \renewcommand\thefootnote{}%
  \renewcommand\@makefntext[1]{\noindent##1}%
  \footnote{#1}%
  \addtocounter{footnote}{-1}%
  \endgroup
}
\makeatother

\definecolor{cverbbg}{gray}{0.90}

\usepackage[stable]{footmisc}

\title{Reconstructing Gene Expression Fields from Histology}
\author[1]{Yujia Cai$^{\circ}$}
\author[1]{Qin Ren$^{\circ}$}
\author[2]{Aosong Feng}
\author[1]{Lixuan Guo}
\author[3]{Yuxin Li}
\author[3]{Bo Chen}
\author[1]{Chenyu You}

\affiliation[1]{Stony Brook University}
\affiliation[2]{Yale University}
\affiliation[3]{Xidian University}
\contribution{$^\circ$Equal contribution}

\metadata[Github]{\url{https://github.com/Y-Research-SBU/Carta}}
\metadata[Correspondence]{\sffamily Chenyu You: \url{chenyu.you@stonybrook.edu}}

\date{\sffamily\today}

\abstract{
Spatial transcriptomics links gene expression to tissue morphology, but sequencing-based assays remain too costly and labor-intensive for routine profiling at scale. This has motivated a growing line of work on inferring spatial expression from standard hematoxylin-and-eosin (H\&E) histology. 
Current methods, however, remain largely tied to assay-indexed prediction:
they estimate expression at measured locations, with spatial context primarily
used to inform predictions defined at those locations. 
This view is misaligned with how molecular signal is organized in tissue. Gene expression is not a set of isolated measurements, but a morphology-conditioned expression field shaped by regional tissue structure and resolved through local transcriptional variation. We formulate histology-based spatial gene expression prediction as \textit{field reconstruction}. \textbf{Carta} reconstructs a morphology-conditioned expression field over the measured spatial coordinates of a spatial transcriptomics assay. It first estimates regional expression states from histology and then refines local expression by propagating information across nearby measured locations. To move beyond losses evaluated independently at sampled coordinates, Carta employs a bidirectional conditional transport objective to shape gene-expression distributions across the reconstruction hierarchy, improving the preservation of sparse expression signals and heterogeneity. Across \textbf{16} organ-specific prediction tasks --- \textbf{10} based on HEST-1k and \textbf{6} constructed from STImage-1K4M, Carta improves average PCC by 12.9 and 2.3 absolute PCC points over the UNI pathology foundation model baseline and the strongest baseline.  
It is particularly effective for sparse genes and spatially localized
expression patterns, whose weak but structured variations can be
under-represented by coordinate-level regression. These results show that morphology-conditioned field reconstruction provides a principled formulation for spatial expression prediction when direct molecular profiling is impractical.
}

\begin{document}
\maketitle

\begingroup

\endgroup
\section{Introduction} \label{sec:intro}
Spatial transcriptomics (ST) measures gene expression while preserving the spatial organization of tissue~\citep{staahl2016visualization,xiao2021tumor,zeng2022spatial,jia2024thitogene}. By linking molecular activity to histological structure, ST has become a powerful tool for studying developmental programs, tumor microenvironments, and disease progression~\citep{van2018gene,wu2024idmir,li2024spadit}. Yet sequencing-based ST assays remain expensive, labor-intensive, and difficult to deploy routinely at scale~\citep{shi2024next,williams2022introduction,staahl2016visualization}. Hematoxylin-and-eosin (H\&E) stained histology, in contrast, is inexpensive, standardized, and widely available in both research and clinical workflows. This gap has motivated a growing line of work on inferring spatial gene expression directly from histology images.

Existing methods approach this problem from several directions. Early approaches learn supervised mappings from local image patches to expression values at measured locations~\citep{he2020integrating,monjo2022efficient}. Later methods incorporate neighborhood graphs~\citep{zeng2022spatial,mejia2023sepal,yang2024spatial,wang2025fmh2st,niu2026ph2st,sun2025bridging,ganguly2025merge} or multi-resolution whole-slide features~\citep{chung2024accurate} to capture broader morphology. Retrieval-based methods estimate expression by aggregating profiles from visually similar reference regions~\citep{xie2023spatially,shi2024spatial,min2024multimodal,chen2025visual,wu2025danet}, while generative approaches based on diffusion~\citep{zhu2025diffusion,liu2025hadm} or flow matching~\citep{huang2025scalable} model conditional expression distributions given histology. Together, these studies show that tissue morphology carries recoverable molecular signal, and that spatial information is essential for extracting it.

\begin{figure}
\centering
\includegraphics[width=1.0\linewidth]{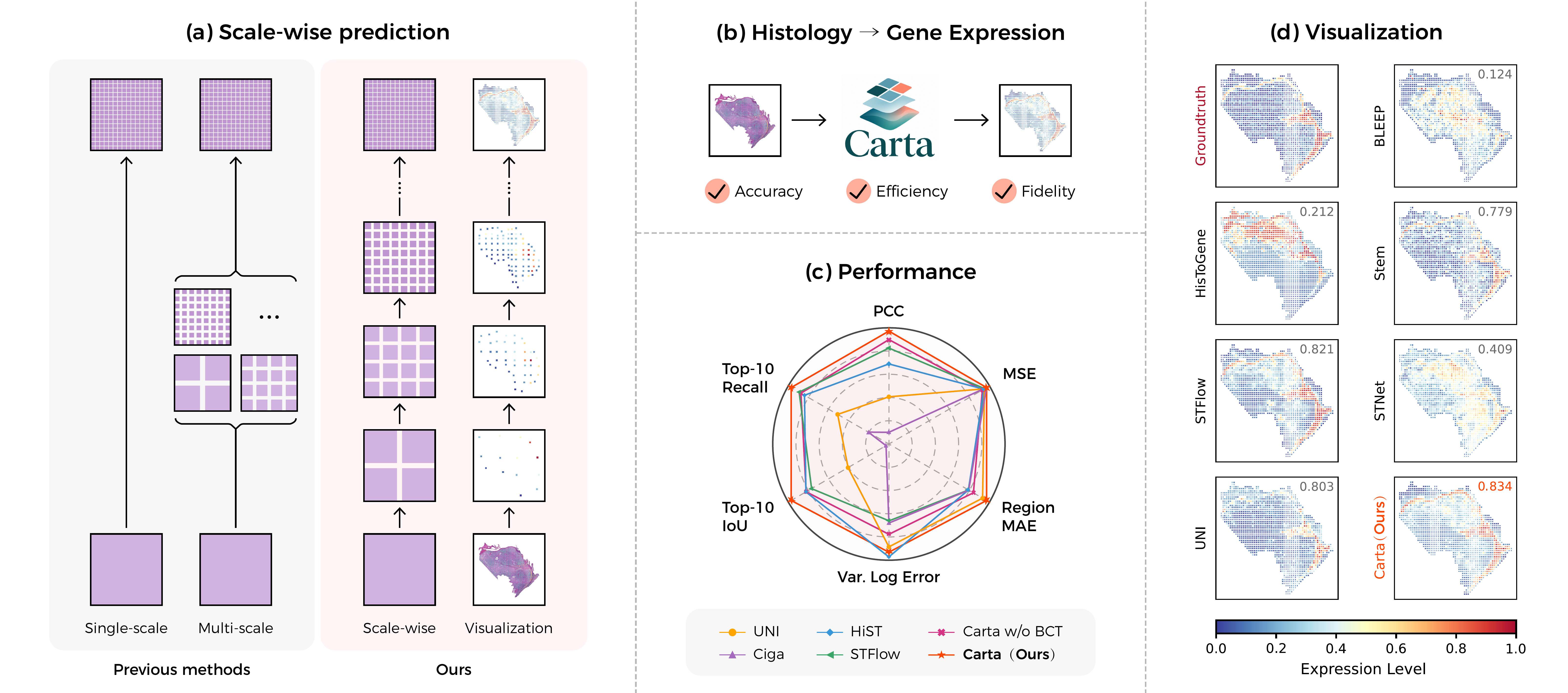}
\caption{\textbf{Carta enables scale-wise reconstruction of gene expression fields from histology.}
(a) Unlike conventional single-scale prediction or multi-scale feature aggregation, Carta progressively reconstructs gene expression from coarse regional states to fine-scale predictions.
(b) Carta predicts spatial gene expression from histology, aiming to improve prediction accuracy, computational efficiency, and reconstruction fidelity.
(c) Across multiple evaluation metrics, Carta demonstrates favorable performance compared with representative baselines, reflecting improvements in both prediction accuracy and field reconstruction quality. All metrics are rescaled such that larger radii indicate better performance. 
(d) Visual comparisons further illustrate that Carta better recovers gene-specific expression patterns and preserves local expression heterogeneity.
}
\label{fig:concept}
\end{figure}


Despite these architectural differences, the supervised target often remains expression at locations measured by the spatial assay~\citep{he2020integrating,mejia2023sepal,xie2023spatially,pang2021leveraging,monjo2022efficient}. 
Even when multiple locations are processed jointly~\citep{zhu2025diffusion,huang2025scalable,chung2024accurate}, the loss is usually evaluated at the measured coordinates. The task is therefore still anchored to the assay's sampled locations. Gene expression in tissue is organized differently~\citep{hu2026histoprism}: it varies as a morphology-conditioned field, shaped by regional tissue structure and resolved through local cellular neighborhoods and transcriptional variation.

Coordinate-level losses supervise the field only through values at sampled
locations. They therefore do not explicitly distinguish between two types of errors:
errors that change individual expression vectors and errors that distort
the spatial organization of the field
\citep{zhu2025diffusion,huang2025scalable,he2020integrating}.
A model can reduce average error at the measured locations while attenuating
sparse genes, blurring boundaries between tissue regions, and washing out
molecular niches spread across nearby locations
\citep{zhu2025diffusion,xie2023spatially,mathieu2015deep}.
For sparse expression profiles, weak or localized deviations may contribute
only marginally to the aggregate coordinate-level objective relative to
larger errors elsewhere in the field.
Consequently, fitting sampled values alone does not ensure that the spatial
organization of the reconstructed field is faithfully preserved~\citep{ledig2017photo,zhao2016loss}. 
This limitation becomes particularly relevant
at finer spatial scales, where increasingly localized expression variation
must be resolved.

We therefore formulate spatial gene expression prediction from histology as a \textit{field reconstruction}. We view the measured ST expression profiles as irregularly spaced samples of an underlying morphology-conditioned expression field. Our goal is to \textit{reconstruct this field across the assayed locations, capturing both regional expression patterns and local variation within each region}.

In this paper, we introduce \textbf{Carta}, which instantiates this field-reconstruction view through a regional-to-local prediction factorization, as illustrated in Figure~\ref{fig:concept}. Carta first infers regional expression states from tissue morphology and uses them to condition progressively finer expression predictions. 
It then refines local expression by propagating information among nearby locations, allowing broad tissue context to inform fine-grained variation. 
To reduce field-level distortions left by coordinate-level losses, Carta uses a bidirectional conditional transport objective~\citep{tanwisuth2021prototype} that shapes gene-expression distributions across the reconstruction hierarchy, improving the preservation of sparse expression signals and heterogeneity.
Across {16} organ-specific prediction tasks, \textit{i.e.}, {10} based on HEST~\citep{jaume2024hest} and {6} constructed from STImage-1K4M~\citep{chen2024stimage}, Carta consistently improves histology-based spatial gene expression prediction over strong baselines. 
It is particularly effective for sparse genes and spatially localized expression patterns, for which weak but structured variations can be under-represented by coordinate-level regression. 
Together, these results show that morphology-conditioned field reconstruction provides a principled formulation for spatial expression prediction when direct molecular profiling is impractical.
Our contributions are as follows:
\begin{itemize}
\item We formulate histology-based spatial gene expression prediction as explicit morphology-conditioned field reconstruction, shifting the modeling perspective and prediction factorization from assay-indexed measurements to an underlying molecular field sampled on irregular spatial layouts.
\item We introduce \textbf{Carta}, which predicts regional expression states and propagates them to condition progressively finer expression estimates over the observed sampling geometry.
\item We develop a bidirectional conditional transport objective for shaping the
distributional structure of the reconstructed field, encouraging the reconstructed landscape to
preserve both sparse expression signals and local heterogeneity beyond
coordinate-level prediction error.
\item We demonstrate consistent improvements over strong baselines across 16 organ-specific prediction tasks, tissues, and sequencing platforms, with especially large gains on sparse genes and spatially localized expression patterns.
\end{itemize}

\section{Related Work} 
\label{sec:related}
\myparagraph{Spatial gene expression prediction from histology images.}
Histology-based spatial gene expression prediction has been studied along several directions~\citep{huang2025scalable}. Early methods predict expression from local tissue morphology, typically by extracting image patches around measured locations and training supervised predictors on the corresponding expression profiles~\citep{xie2023spatially,he2020integrating}. 
Other methods expand the visual field of view with neighboring locations, graph structure, slide-level representations, or multi-resolution whole-slide features, allowing predictions to use broader tissue context~\citep{pang2021leveraging,chung2024accurate,ganguly2025merge,wang2025fmh2st}. 
Recent hierarchical approaches further explore multi-scale latent feature propagation through encoder-decoder architectures~\citep{wu2026hist} and next-scale autoregressive prediction along the gene dimension~\citep{ouyang2026genar}. 
More recent work has introduced retrieval-based estimation from visually similar reference regions and generative modeling through diffusion or flow-based conditional distributions~\citep{huang2025scalable,pang2021leveraging,zhu2025diffusion,he2020integrating,xie2023spatially}.

These methods differ in architecture and learning objective, but they are usually trained around the same observed quantity: expression values at locations measured by the assay. Spatial information is incorporated through patches, graphs, slide context, retrieval sets, or generative priors, yet the target remains a set of sampled measurements~\citep{zhu2025diffusion,xie2023spatially,he2020integrating}. 
In contrast, Carta changes how predictions are
factorized over the sampled geometry. 
Rather than using spatial structure primarily to contextualize predictions on a fixed support, Carta constructs a sequence of coordinate-derived supports, predicts regional expression states, and propagates these predicted states to condition progressively finer layouts.

\begin{figure*}[t]
\centering
\includegraphics[width=1.0\linewidth]{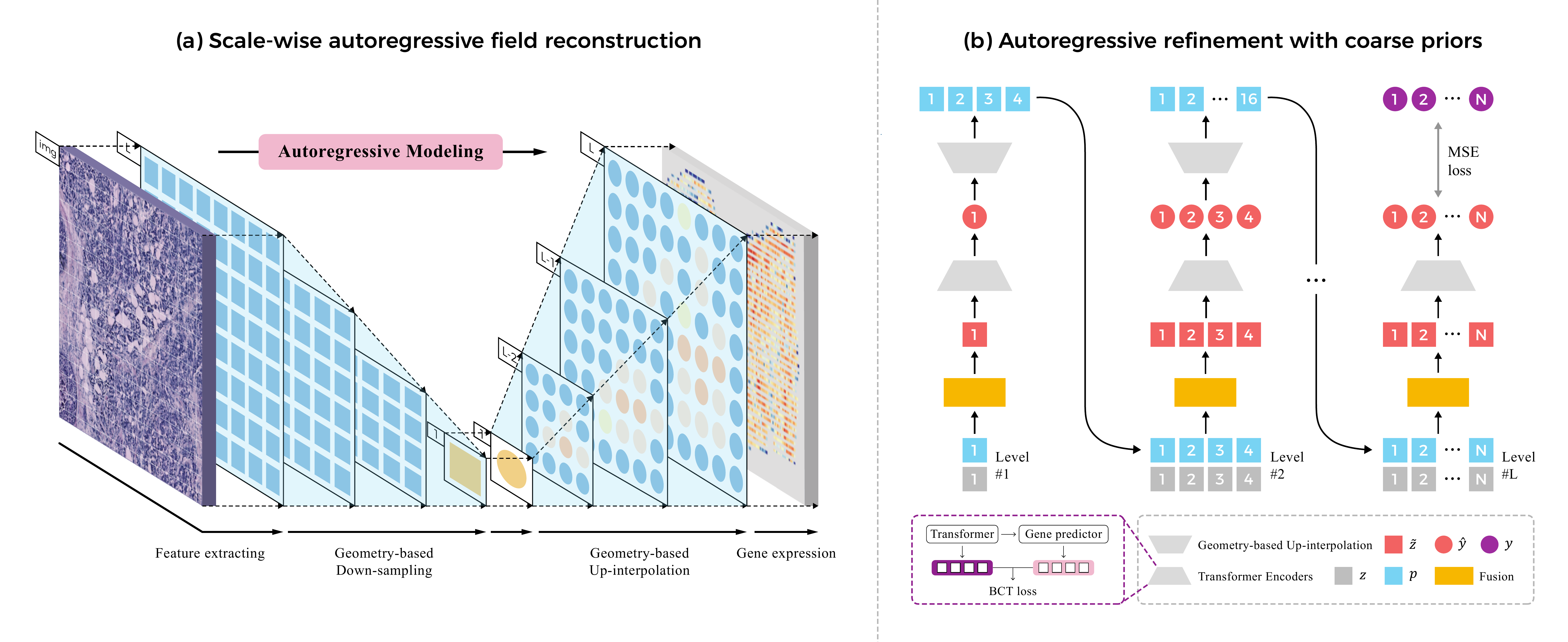}
\vspace{-8pt}
\caption{
\textbf{Overview of Carta.}
Carta builds a spatial hierarchy from measured ST coordinates and reconstructs expression from regional states to local predictions.
At each scale, propagated expression states condition local histology features; the final prediction is supervised at the measured coordinates, while BCT shapes the distributional structure of the reconstructed field.
}
\label{fig:framework}
\end{figure*}

\myparagraph{Hierarchical autoregressive generation in vision.}
Autoregressive visual models factorize image generation into a sequence of conditional predictions~\citep{tian2024visual,sun2024autoregressive,esser2021taming}. Early methods often operate on discrete visual tokens and generate them under a fixed order~\citep{esser2021taming,yu2022scaling}. Recent work moves beyond a single flat sequence by introducing multi-scale tokenizations or scale-wise prediction orders, so that global layout is modeled before local texture and detail~\citep{tian2024visual,ren2024flowar}. In this line of work, the ordering is part of the model: it determines which information is available when later predictions are made. 

In Carta, the ordering is defined by a coordinate-derived spatial hierarchy rather than image resolution. 
Spatial gene expression is organized across tissue regions and local cellular
neighborhoods~\citep{zhu2025diffusion,williams2022introduction}. Carta first predicts coarse regional gene-expression states from histology and propagates these states across the spatial hierarchy to condition progressively finer gene-expression estimates at measured locations. This regional-to-local factorization enables local predictions to incorporate regional molecular context while preserving fine-scale spatial variation.

\myparagraph{Conditional transport and distributional structure.}
Existing histology-based expression prediction methods are trained with paired regression losses, since each measured location has a corresponding expression vector. This supervision is natural, but it only compares predictions and observations at matched samples. At the level of a \textit{reconstructed field}, what remains insufficiently characterized is how predicted profiles are distributed within and across tissue regions. A model can fit many paired profiles while under-representing the tails of the expression distribution, especially for sparse genes, low-expression programs, or region-specific heterogeneity~\citep{bishop1994mixture,guilhoto2026multimodal,zhu2025diffusion}.

Conditional transport models directional conditional assignments between
two sets, providing an efficient alternative to a balanced transport coupling with prescribed shared marginals~\citep{li2023patchct,tanwisuth2021prototype,zheng2021exploiting}. In histology-based spatial expression prediction, however, transport objectives have rarely been used to model the structure of the reconstructed expression field~\citep{ren2025otsurv,feng2024tarot,halmos2025hierarchical,liu2019wasserstein}. 
Motivated by this, we adapt the bidirectional conditional transport construction to shape gene-expression distributions across the prediction hierarchy.
Directly applying transport to intermediate-scale expression profiles, however, would require constructing pseudo-labels for regional nodes.
Since these nodes are derived from coordinate-based partitioning of irregularly sampled locations rather than directly measured, spatially aggregated expression profiles may provide unreliable transport targets for intermediate states.
Instead, we relate node prediction features to gene-specific anchors defined by the shared expression decoder, avoiding the need for intermediate-scale expression labels. 
BCT is applied across the prediction
hierarchy, acting on regional-node features at coarse scales and spot-level features at the finest scale. This gene-anchored transport objective complements coordinate-level regression and provides larger gains for sparse genes while improving field-level reconstruction quality.
\section{Method}
\label{sec:method}
A spatial transcriptomics assay measures gene expression at a finite, irregular set of spatial locations. For histology-based prediction, each measured coordinate is paired with an H\&E patch centered at the same location. We model these paired observations as samples from a morphology-conditioned expression field, rather than as independent labels attached to spatial coordinates.

Carta estimates this field by separating regional state estimation from local prediction. At coarse scales, regional nodes aggregate morphology over nearby measured locations and produce expression states that capture broad tissue context. At finer scales, predictions at measured coordinates are conditioned on these regional states and updated using neighboring locations. Training uses coordinate-level regression for the measured expression values, together with a bidirectional conditional transport objective that shapes the distributional structure of the reconstructed field. The overall framework is illustrated in Figure~\ref{fig:framework}. Additional method details are provided in Supplementary Material~\ref{app:method-derivations}.   We next define the field reconstruction problem, describe the regional-to-local predictor, and introduce the bidirectional conditional transport objective.

\subsection{Field Reconstruction Formulation}
Consider a whole-slide histology image paired with a spatial transcriptomics sample. 
Let $\mathbf{C}=\{c_i\}_{i=1}^{N}$ denote the measured spatial coordinates, with $c_i\in\mathbb{R}^{2}$, and let $\mathbf{Y}=\{y_i\}_{i=1}^{N}$ denote the corresponding expression profiles, with $y_i\in\mathbb{R}^{G}$ for $G$ target genes. 
For each coordinate $c_i$, we extract a local H\&E patch $x_i$ and write $\mathbf{X}=\{x_i\}_{i=1}^{N}$.

We model these observations as samples from a morphology-conditioned expression field,
\begin{equation}
    \mathcal{F}_{\mathbf{X}}: \Omega \rightarrow \mathbb{R}^{G},
\end{equation}
where $\Omega \subset \mathbb{R}^{2}$ is the tissue domain and 
$y_i=\mathcal{F}_{\mathbf{X}}(c_i)$ is the measured field value. 
Given the patch set and measured coordinates, Carta estimates field values at the assayed locations:
\begin{equation}
    \hat{\mathbf{Y}}
    =
    \hat{\mathcal F}_{\theta}(\mathbf{X},\mathbf{C}),
    \qquad
    \hat y_i
    =
    \hat{\mathcal F}_{\theta}(\mathbf{X},\mathbf{C})_i .
\end{equation}
This highlights that prediction at one location may depend on the full histology-coordinate set, not only on the local patch.
Carta represents the field on a sequence of coarse-to-fine spatial layouts
$\{\mathcal{V}^{(1)},\ldots,\mathcal{V}^{(L)}\}$, where $\mathcal{V}^{(1)}$ contains regional nodes and $\mathcal{V}^{(L)}$ corresponds to the original measured coordinates. 
The model produces estimates:
$\{\hat{\mathbf{Y}}^{(1)},\ldots,\hat{\mathbf{Y}}^{(L)}\}$, with $\hat{\mathbf{Y}}^{(L)}=\hat{\mathbf{Y}}$.

\subsection{Regional-to-Local Reconstruction}
\label{sec:regional-local}
Carta builds the spatial hierarchy from the measured coordinates rather than from an image grid. 
We encode each H\&E patch with a pathology foundation model (PFM)~\citep{ciga2022self,xu2024whole,chen2024towards}:
\begin{equation}
    h_i = f_{\mathrm{PFM}}(x_i),
    \qquad h_i\in\mathbb{R}^{d_h}.
\end{equation}
For the finest scale, each node corresponds to one measured ST location, so 
$\mathcal{V}^{(L)}=\{(h_i,c_i)\}_{i=1}^{N}$. 
For each coarser scale $l<L$, we partition the normalized coordinate frame into bins and keep only bins containing measured locations. 
The $j$-th retained bin defines a regional node $R_j^{(l)}$, with feature and coordinate:
\begin{equation}
    h_j^{(l)}
    =
    \frac{1}{|R_j^{(l)}|}
    \sum_{i\in R_j^{(l)}} h_i,
    \qquad
    c_j^{(l)}
    =
    \frac{1}{|R_j^{(l)}|}
    \sum_{i\in R_j^{(l)}} c_i .
\end{equation}
All distances are computed in the normalized coordinate frame; we keep the notation $c_i$ for simplicity. 
Additional binning details are given in Supplementary Material~\ref{app:coordinate-binning}.
For $l\geq 2$, expression states are transferred from scale $l-1$ to scale $l$ by distance-weighted interpolation. 
For node $i$ at scale $l$, let $\mathcal{N}_i^{(l)}$ denote its nearest nodes from scale $l-1$, with $|\mathcal{N}_i^{(l)}|=\min(K,N_{l-1})$; if fewer than $K$ valid nodes exist, all of them are used. 
The interpolated expression state is:
\begin{equation}
    p_i^{(l)}
    =
    \sum_{j\in\mathcal{N}_i^{(l)}}
    w_{ij}^{(l)} \hat y_j^{(l-1)},
\end{equation}
where,
\begin{equation}
    w_{ij}^{(l)}
    =
    \frac{
    \exp\!\left(-\|c_i^{(l)}-c_j^{(l-1)}\|_2^2/\lambda_u\right)
    }{
    \sum_{j'\in\mathcal{N}_i^{(l)}}
    \exp\!\left(-\|c_i^{(l)}-c_{j'}^{(l-1)}\|_2^2/\lambda_u\right)
    }.
\end{equation}
Here $\lambda_u>0$ is the interpolation bandwidth.
At the coarsest scale, Carta initializes the expression condition with a learnable gene-wise Gaussian prior:
\begin{equation}
    p^{(1)}
    =
    \mu+\exp(\log\sigma)\odot\varepsilon,
    \qquad
    \varepsilon\sim\mathcal{N}(0,I),
\end{equation}
where $\mu,\log\sigma\in\mathbb{R}^{G}$. 
For every coarsest-scale node $i$, we set $p_i^{(1)}=p^{(1)}$ during training and use $p_i^{(1)}=\mu$ at evaluation unless otherwise specified. Further details of cross-scale expression-state propagation and coarsest-scale initialization are provided in Supplementary Material~\ref{app:cross-scale}.

\myparagraph{Conditional prediction at each scale.}
At scale $l$, Carta predicts expression from the histology representation at the current node and the expression condition $p_i^{(l)}$. 
We project both into the model latent space:
\begin{equation}
    z_i^{(l)} = W_h h_i^{(l)}\in\mathbb{R}^{d},
    \qquad
    e_i^{(l)} = W_p p_i^{(l)}\in\mathbb{R}^{d}.
\end{equation}
Let $\mathcal{M}_i^{(l)}$ denote the within-scale neighborhood of node $i$.
We define the query, key, and value for cross-attention as
\begin{equation}
    Q_i^{(l)} = z_i^{(l)}, \qquad
    K_i^{(l)} = V_i^{(l)}
    = \{e_j^{(l)}:j\in\mathcal{M}_i^{(l)}\}.
\end{equation}
Carta conditions the histology state on nearby expression conditions through cross-attention:
\begin{equation}
    \tilde z_i^{(l)}
    =
    z_i^{(l)}
    +
    \operatorname{CrossAttn}
    \left(
        Q_i^{(l)}, K_i^{(l)}, V_i^{(l)}
    \right).
\end{equation}
A Transformer encoder then produces the contextualized node representation
\begin{equation}
    u_i^{(l)}
    =
    \mathrm{Transformer}
    \big(\{\tilde z_j^{(l)}:j\in\mathcal{M}_i^{(l)}\}\big)_i .
\end{equation}
The attention operator incorporates geometric information~\citep{huang2025scalable}, with its implementation detailed in Supplementary Material~\ref{app:geo-attention}.
A two-layer prediction head maps $u_i^{(l)}$ to expression values. 
We obtain an intermediate prediction feature
$f_i^{(l)}=\phi_1(u_i^{(l)})\in\mathbb{R}^{d_f}$
and predict expression by:
\begin{equation}
    \hat y_i^{(l)}
    =
    \psi\!\left(W_of_i^{(l)}+b_o\right),
\end{equation}
where $\phi_1(\cdot)$ is the intermediate projection, $W_o\in\mathbb{R}^{G\times d_f}$, and $b_o\in\mathbb{R}^{G}$. 
The activation $\psi$ is $\mathrm{softplus}$ for non-negative expression targets and the identity map for standardized targets. 
Predictions at intermediate scales are passed to the next scale by the interpolation operator above. 
At the finest scale, the nodes are the original measured coordinates, and $\hat{\mathbf{Y}}^{(L)}$ is the final reconstructed field.

\subsection{Bidirectional Conditional Transport Objective}
\label{sec:bct}

Coordinate-level regression matches each predicted expression vector to the measured vector at the same location, but it does not explicitly characterize how gene-specific responses are distributed across the reconstructed field. 
Carta therefore adds a Bidirectional Conditional Transport (BCT) objective~\citep{tanwisuth2021prototype}, which relates node prediction features to gene anchors defined by the output layer.

Let $\mathbf{F}^{(l)}=\{f_i^{(l)}\}_{i=1}^{N^{(l)}}$ denote the node prediction features at scale $l$, where $N^{(l)}=|\mathcal{V}^{(l)}|$. 
We use the rows of the shared output projection $W_o$ as gene anchors: 
\[
    \boldsymbol{g}
    =
    \{g_k\}_{k=1}^{G},
    \qquad
    g_k=\operatorname{sg}\!\left(W_o[k,:]\right).
\]
where $\operatorname{sg}(\cdot)$ denotes stop-gradient within BCT; $W_o$ remains trainable through the expression regression objective. The similarity between gene anchor $k$ and node feature $i$ is:
$S_{ki}^{(l)}={g_k}^\top f_i^{(l)}$,
and the transport cost is the cosine distance:
\begin{equation}
    D_{ki}^{(l)}
    =
    1-
    \frac{{g_k}^\top f_i^{(l)}}
    {\|g_k\|_2\|f_i^{(l)}\|_2}.
\end{equation}
To account for uneven gene activity in a minibatch, BCT maintains a dynamic prior 
$\mathbf q\in\Delta^{G-1}$ over gene anchors, where
$\Delta^{G-1}=\{\mathbf q\in\mathbb{R}_{+}^{G}:\sum_{k=1}^{G}q_k=1\}$.
The prior is initialized uniformly and updated online from batch matching statistics; details are provided in Supplementary Material~\ref{app:bct-anchor-prior}. 
Given $\mathbf q$, the node-to-anchor transport weights are:
\begin{equation}
    \pi_{ki}^{s,(l)}
    =
    \mathrm{softmax}_{k}\!\left(
    \frac{S_{ki}^{(l)}}{\tau}+\log(q_k+\delta)
    \right),
\end{equation}
where $\tau$ is the temperature and $\delta>0$ is used for numerical stability. These weights represent soft matching between spatial nodes and gene anchors. The node-to-anchor loss is:
\begin{equation}
    \mathcal{L}_{s}^{(l)}
    =
    \frac{\alpha}{N^{(l)}}
    \sum_{i=1}^{N^{(l)}}\sum_{k=1}^{G}
    D_{ki}^{(l)}\pi_{ki}^{s,(l)}.
\end{equation}
The reverse transport weights assign each gene anchor to spatial nodes:
\begin{equation}
    \pi_{ki}^{t,(l)}
    =
    \mathrm{softmax}_{i}\!\left(
    \frac{S_{ki}^{(l)}}{\tau}
    \right),
\end{equation}
with loss,
\begin{equation}
    \mathcal{L}_{t}^{(l)}
    =
    (1-\alpha)
    \sum_{k=1}^{G}q_k
    \sum_{i=1}^{N^{(l)}}
    D_{ki}^{(l)}\pi_{ki}^{t,(l)} .
\end{equation}
Here $\alpha\in[0,1]$ balances the two transport directions.
The BCT loss at scale $l$ is
$\mathcal{L}_{\mathrm{BCT}}^{(l)} = \mathcal{L}_{s}^{(l)}+\mathcal{L}_{t}^{(l)}$,
and the multi-scale regularizer is: $\mathcal{L}_{\mathrm{BCT}} = \frac{1}{L}\sum_{l=1}^{L}\mathcal{L}_{\mathrm{BCT}}^{(l)}$.
Regression supervision is applied at the finest scale:
$\mathcal{L}_{\mathrm{reg}} = \mathrm{MSE}(\hat{\mathbf{Y}}^{(L)},\mathbf{Y})$.
The final training objective is:
\begin{equation}
    \mathcal{L}
    =
    \mathcal{L}_{\mathrm{reg}}
    +
    \lambda_{\mathrm{BCT}}\mathcal{L}_{\mathrm{BCT}},
\end{equation}
where $\lambda_{\mathrm{BCT}}$ controls the strength of the transport objective.

\section{Experiments} 
\label{sec:exp}
In this section, we evaluate Carta on spatial gene expression prediction from H\&E histology.
Our experiments cover two paired histology--spatial-transcriptomics resources and compare Carta against methods that differ in visual representation, use of reference data, spatial context modeling, and generative modeling. Additional analyses in Supplementary Material~\ref{app:additional-experiments} examine the predictive capacity of frozen pathology features (~\ref{app:pathology-feature-sufficiency}).
We then evaluate prediction at held-out spatial coordinates
(~\ref{app:heldout-coordinates}) and spatial field diagnostics
(~\ref{app:field-level-analysis}). Further analyses examine scale-schedule
selection (~\ref{app:scale-schedule}), cross-scale fusion
(~\ref{app:cross-fusion-strategy-ablation}), coarsest-scale
initialization (~\ref{app:initialization}), scale-wise supervision
(~\ref{app:scale-mse-ablation}), BCT hyperparameter
sensitivity (~\ref{app:bct-sensitivity}), gene-anchor design
(~\ref{app:bct-anchor-ablation}), and the relationship between gene-anchor
geometry and gene co-expression (~\ref{app:anchor-coexpression}).
Finally, we report computational efficiency
(~\ref{app:efficiency}), with qualitative analyses in Supplementary Material~\ref{app:qualitative-analysis}.

\begin{table}[!t]
\centering
\begin{minipage}{\linewidth}
\captionsetup{skip=5pt}
\caption{
\textbf{Organ-wise comparison on HEST-1k.}
Methods are arranged in rows, and organs are arranged in columns.
Each entry reports the mean PCC, with subscripts indicating the
standard deviation across multiple random seeds.
The best results are highlighted in bold, and the second-best
results are underlined. Avg. denotes the average across organs.
}
\label{tab:hest-organ-wise}
\centering
\scriptsize
\setlength{\tabcolsep}{4.5pt}
\renewcommand{\arraystretch}{1.08}
\begin{adjustbox}{max width=\linewidth}
\begin{tabular}{lccccccccccc}
\toprule
\textbf{Method}
& \textbf{CCRCC}
& \textbf{COAD}
& \textbf{HCC}
& \textbf{IDC}
& \textbf{LUNG}
& \textbf{LYMPH}
& \textbf{PAAD}
& \textbf{PRAD}
& \textbf{READ}
& \textbf{SKCM}
& \textbf{Avg.} \\
\midrule

\rowcolor{hestbg}
Ciga
& $\res{0.126}{.001}$
& $\res{0.250}{.000}$
& $\res{0.043}{.001}$
& $\res{0.405}{.000}$
& $\res{0.514}{.000}$
& $\res{0.217}{.001}$
& $\res{0.396}{.000}$
& $\res{0.331}{.001}$
& $\res{0.046}{.000}$
& $\res{0.484}{.000}$
& $0.281$ \\

\rowcolor{hestbg}
Gigapath
& $\res{0.182}{.001}$
& $\res{0.261}{.003}$
& $\res{0.054}{.000}$
& $\res{0.491}{.001}$
& $\res{0.546}{.002}$
& $\res{0.248}{.001}$
& $\res{0.426}{.000}$
& $\res{0.370}{.000}$
& $\res{0.155}{.001}$
& $\res{0.540}{.002}$
& $0.327$ \\

\rowcolor{hestbg}
UNI
& $\res{0.186}{.000}$
& $\res{0.295}{.001}$
& $\res{0.052}{.001}$
& $\res{0.487}{.002}$
& $\res{0.512}{.001}$
& $\res{0.233}{.000}$
& $\res{0.423}{.000}$
& $\res{0.357}{.002}$
& $\res{0.163}{.000}$
& $\res{0.613}{.001}$
& $0.332$ \\

\rowcolor{hestbg}
BLEEP
& $\res{0.286}{.001}$
& $\res{0.289}{.004}$
& $\res{0.057}{.006}$
& $\res{0.521}{.005}$
& $\res{0.597}{.002}$
& $\res{0.221}{.002}$
& $\res{0.407}{.008}$
& $\res{0.369}{.005}$
& $\res{0.236}{.024}$
& $\res{0.518}{.008}$
& $0.350$ \\

\rowcolor{hestbg}
DANet
& $\res{0.251}{.002}$
& $\res{0.225}{.003}$
& $\res{0.061}{.003}$
& $\res{0.538}{.001}$
& $\res{0.545}{.003}$
& $\res{0.255}{.002}$
& $\res{0.387}{.002}$
& $\res{0.384}{.004}$
& $\res{0.205}{.022}$
& $\res{0.611}{.001}$
& $0.346$ \\

\rowcolor{hestbg}
HisToGene
& $\res{0.155}{.011}$
& $\res{0.215}{.019}$
& $\res{0.038}{.007}$
& $\res{0.311}{.012}$
& $\res{0.412}{.005}$
& $\res{0.242}{.013}$
& $\res{0.234}{.011}$
& $\res{0.258}{.009}$
& $\res{0.031}{.018}$
& $\res{0.233}{.008}$
& $0.213$ \\

\rowcolor{hestbg}
Stem
& $\res{0.242}{.019}$
& $\res{0.309}{.006}$
& $\res{0.064}{.007}$
& $\res{0.517}{.022}$
& $\res{0.572}{.012}$
& $\res{0.238}{.018}$
& $\res{0.420}{.012}$
& $\res{0.350}{.010}$
& $\res{0.227}{.017}$
& $\res{0.652}{.005}$
& $0.359$ \\

\rowcolor{hestbg}
STFlow
& $\second{0.329}{.022}$
& $\second{0.327}{.017}$
& $\second{0.121}{.010}$
& $\res{0.547}{.020}$
& $\res{0.606}{.006}$
& $\res{0.260}{.007}$
& $\res{0.486}{.009}$
& $\second{0.400}{.008}$
& $\second{0.240}{.008}$
& $\res{0.695}{.004}$
& $\underline{0.401}$ \\

\rowcolor{hestbg}
STNet
& $\res{0.166}{.006}$
& $\res{0.276}{.003}$
& $\res{0.097}{.006}$
& $\res{0.407}{.006}$
& $\res{0.566}{.003}$
& $\res{0.264}{.002}$
& $\res{0.427}{.009}$
& $\res{0.372}{.005}$
& $\res{0.158}{.029}$
& $\res{0.439}{.010}$
& $0.317$ \\

\rowcolor{hestbg}
TRIPLEX
& $\res{0.284}{.004}$
& $\res{0.302}{.002}$
& $\res{0.056}{.005}$
& $\best{0.599}{.003}$
& $\res{0.605}{.002}$
& $\second{0.301}{.002}$
& $\second{0.487}{.005}$
& $\res{0.398}{.006}$
& $\res{0.203}{.011}$
& $\res{0.694}{.006}$
& $0.393$ \\

\rowcolor{hestbg}
FmH2ST
& $\res{0.229}{.008}$
& $\res{0.267}{.006}$
& $\res{0.081}{.004}$
& $\res{0.541}{.011}$
& $\res{0.562}{.008}$
& $\res{0.264}{.004}$
& $\res{0.445}{.009}$
& $\res{0.351}{.003}$
& $\res{0.192}{.014}$
& $\res{0.656}{.003}$
& $0.359$ \\

\rowcolor{hestbg}
MERGE
& $\res{0.262}{.009}$
& $\res{0.307}{.007}$
& $\res{0.097}{.002}$
& $\res{0.586}{.006}$
& $\second{0.608}{.003}$
& $\res{0.282}{.010}$
& $\res{0.458}{.006}$
& $\res{0.387}{.002}$
& $\res{0.235}{.016}$
& $\second{0.696}{.004}$
& $0.392$ \\

\rowcolor{hestbg}
HiST
& $\res{0.273}{.002}$
& $\res{0.292}{.004}$
& $\res{0.084}{.001}$
& $\res{0.591}{.008}$
& $\res{0.579}{.001}$
& $\res{0.248}{.004}$
& $\second{0.487}{.007}$
& $\res{0.371}{.003}$
& $\res{0.176}{.008}$
& $\res{0.684}{.003}$
& $0.379$ \\

\rowcolor{hestbg}
GenAR
& $\res{0.274}{.006}$
& $\res{0.298}{.005}$
& $\res{0.077}{.003}$
& $\res{0.568}{.007}$
& $\res{0.521}{.003}$
& $\res{0.233}{.002}$
& $\res{0.397}{.004}$
& $\res{0.333}{.004}$
& $\res{0.198}{.009}$
& $\res{0.644}{.008}$
& $0.354$ \\

\midrule
\rowcolor{hestbg}
\textbf{Carta}
& $\best{0.335}{.007}$
& $\best{0.363}{.008}$
& $\best{0.132}{.006}$
& $\second{0.595}{.009}$
& $\best{0.621}{.004}$
& $\best{0.315}{.005}$
& $\best{0.516}{.010}$
& $\best{0.422}{.004}$
& $\best{0.245}{.033}$
& $\best{0.703}{.006}$
& $\mathbf{0.425}$ \\

\bottomrule
\end{tabular}
\end{adjustbox}
\vspace{-8pt}
\end{minipage}
\end{table}

\subsection{Setup}

\myparagraph{Datasets.}
We evaluate on two H\&E-paired spatial transcriptomics resources: \texttt{HEST-1k}~\citep{jaume2024hest} and \texttt{STImage-1K4M}~\citep{chen2024stimage}. 
From these resources, we define {16} prediction tasks: {10} cancer-specific tasks from \texttt{HEST-1k} and {6} organ-level tasks constructed from \texttt{STImage-1K4M}. 
Together, these tasks span diverse tissue morphologies, disease contexts, sequencing platforms, and expression patterns.

The \texttt{HEST-1k} tasks include IDC, PRAD, PAAD, SKCM, COAD, READ, CCRCC, HCC, LUNG,
and LYMPH.  
For \texttt{HEST-1k}, the performance is evaluated using the patient-disjoint cross-validation setup of prior work~\citep{huang2025scalable,jaume2024hest}. 
The \texttt{STImage-1K4M} tasks include Brain, Breast, Heart, Liver,
Pancreas, and Skin. For each \texttt{STImage-1K4M} task, we construct
a patient-disjoint 80/20 train/test split at the slide level,
ensuring that slides from the same patient do not appear in
both partitions. For all methods, architecture and hyperparameters
are selected on \texttt{HEST-1k} and then fixed for all \texttt{STImage-1K4M}
tasks, without further hyperparameter tuning. 
For each evaluation split, we apply a per-spot log1p transformation and
select the top 50 highly variable genes using only the training partition. Since spatially localized expression can exhibit substantial variability across spots, the selected HVG panel also includes sparsely expressed genes, enabling further analysis of prediction performance across different sparsity levels. 
The resulting gene set is then fixed for all compared methods on that split.
All methods are evaluated using identical HEST patient-disjoint folds or
STImage patient-disjoint partitions.
The final evaluation contains {800} task-specific gene targets across {16} prediction tasks. 
Further dataset statistics, platform compositions, and evaluation protocol details are provided in Supplementary Material~\ref{app:dataset}.

\Needspace{0.40\textheight}
\begin{wraptable}{r}{0.55\textwidth}
\begin{minipage}{\linewidth}
\centering
\captionsetup{skip=5pt}
\vspace{-12pt}
\caption{
\textbf{Organ-wise comparison on STImage-1K4M.}
Methods are arranged in rows, and organs are arranged in columns.
Each entry reports the mean PCC, with subscripts indicating the
standard deviation across multiple random seeds.
The best results are highlighted in bold, and the second-best
results are underlined. Avg. denotes the average across organs.
}
\label{tab:stimage-organ-wise}
\centering
\scriptsize
\setlength{\tabcolsep}{0.5pt}
\renewcommand{\arraystretch}{1.08}
\begin{adjustbox}{max width=\linewidth}
\begin{tabular}{lccccccc}
\toprule
\textbf{Method}
& \textbf{Brain}
& \textbf{Breast}
& \textbf{Heart}
& \textbf{Liver}
& \textbf{Pancreas}
& \textbf{Skin}
& \textbf{Avg.} \\
\midrule

\rowcolor{stimagebg}
Ciga
& $\res{0.410}{.001}$
& $\res{0.489}{.000}$
& $\res{0.581}{.000}$
& $\res{0.407}{.001}$
& $\res{0.543}{.002}$
& $\res{0.331}{.001}$
& $0.460$ \\

\rowcolor{stimagebg}
Gigapath
& $\res{0.463}{.001}$
& $\res{0.523}{.002}$
& $\res{0.643}{.001}$
& $\res{0.424}{.001}$
& $\res{0.608}{.001}$
& $\res{0.343}{.000}$
& $0.501$ \\

\rowcolor{stimagebg}
UNI
& $\res{0.483}{.002}$
& $\res{0.486}{.001}$
& $\res{0.638}{.002}$
& $\res{0.434}{.000}$
& $\res{0.585}{.001}$
& $\res{0.302}{.002}$
& $0.488$ \\

\rowcolor{stimagebg}
BLEEP
& $\res{0.497}{.006}$
& $\res{0.442}{.007}$
& $\res{0.632}{.005}$
& $\res{0.566}{.006}$
& $\res{0.713}{.011}$
& $\res{0.487}{.012}$
& $0.556$ \\

\rowcolor{stimagebg}
DANet
& $\res{0.535}{.009}$
& $\res{0.433}{.012}$
& $\res{0.629}{.004}$
& $\res{0.572}{.005}$
& $\res{0.754}{.013}$
& $\res{0.450}{.008}$
& $0.562$ \\

\rowcolor{stimagebg}
HisToGene
& $\res{0.232}{.004}$
& $\res{0.391}{.013}$
& $\res{0.562}{.006}$
& $\res{0.543}{.013}$
& $\res{0.511}{.018}$
& $\res{0.351}{.007}$
& $0.432$ \\

\rowcolor{stimagebg}
Stem
& $\res{0.515}{.016}$
& $\res{0.513}{.018}$
& $\res{0.689}{.007}$
& $\res{0.586}{.009}$
& $\res{0.739}{.030}$
& $\res{0.444}{.013}$
& $0.581$ \\

\rowcolor{stimagebg}
STFlow
& $\res{0.568}{.002}$
& $\res{0.535}{.023}$
& $\best{0.768}{.004}$
& $\res{0.615}{.012}$
& $\second{0.798}{.005}$
& $\second{0.508}{.004}$
& $\underline{0.632}$ \\

\rowcolor{stimagebg}
STNet
& $\res{0.474}{.009}$
& $\res{0.454}{.012}$
& $\res{0.693}{.011}$
& $\second{0.618}{.008}$
& $\res{0.623}{.014}$
& $\res{0.424}{.010}$
& $0.548$ \\

\rowcolor{stimagebg}
TRIPLEX
& $\res{0.551}{.007}$
& $\res{0.544}{.014}$
& $\res{0.739}{.005}$
& $\res{0.598}{.016}$
& $\res{0.797}{.009}$
& $\res{0.499}{.006}$
& $0.621$ \\

\rowcolor{stimagebg}
FmH2ST
& $\res{0.536}{.010}$
& $\res{0.509}{.002}$
& $\res{0.697}{.008}$
& $\res{0.574}{.019}$
& $\res{0.752}{.026}$
& $\res{0.450}{.009}$
& $0.586$ \\

\rowcolor{stimagebg}
MERGE
& $\res{0.546}{.003}$
& $\res{0.553}{.009}$
& $\res{0.736}{.002}$
& $\res{0.595}{.013}$
& $\res{0.795}{.017}$
& $\res{0.486}{.004}$
& $0.619$ \\

\rowcolor{stimagebg}
HiST
& $\second{0.583}{.001}$
& $\second{0.579}{.005}$
& $\res{0.744}{.003}$
& $\res{0.517}{.006}$
& $\res{0.594}{.013}$
& $\res{0.470}{.002}$
& $0.581$ \\

\rowcolor{stimagebg}
GenAR
& $\res{0.512}{.003}$
& $\res{0.485}{.007}$
& $\res{0.713}{.004}$
& $\res{0.560}{.014}$
& $\res{0.765}{.012}$
& $\res{0.462}{.005}$
& $0.583$ \\

\midrule
\rowcolor{stimagebg}
\textbf{Carta}
& $\best{0.586}{.016}$
& $\best{0.595}{.013}$
& $\second{0.763}{.006}$
& $\best{0.659}{.022}$
& $\best{0.806}{.017}$
& $\best{0.509}{.005}$
& $\mathbf{0.653}$ \\

\bottomrule
\end{tabular}
\end{adjustbox}
\par
\end{minipage}
\end{wraptable}

\myparagraph{Baselines.}
We compare Carta with fourteen baselines: Ciga~\citep{ciga2022self}, Gigapath~\citep{xu2024whole}, UNI~\citep{chen2024towards}, BLEEP~\citep{xie2023spatially}, DANet~\citep{wu2025danet}, HisToGene~\citep{pang2021leveraging}, Stem~\citep{zhu2025diffusion}, STFlow~\citep{huang2025scalable}, STNet~\citep{he2020integrating}, TRIPLEX~\citep{chung2024accurate}, FmH2ST~\citep{wang2025fmh2st}, MERGE~\citep{ganguly2025merge}, HiST~\citep{wu2026hist}, and GenAR~\citep{ouyang2026genar}. 
These baselines cover six modeling families:
pathology-foundation-model feature regression
~\citep{ciga2022self,xu2024whole,chen2024towards},
image--expression alignment
~\citep{xie2023spatially,wu2025danet},
task-specific neural regression models
~\citep{pang2021leveraging,he2020integrating},
generative expression modeling
~\citep{zhu2025diffusion,huang2025scalable},
spatial-context modeling
~\citep{chung2024accurate,wang2025fmh2st,ganguly2025merge},
and hierarchical prediction
~\citep{wu2026hist,ouyang2026genar}.

\Needspace{3\baselineskip}
For Ciga, Gigapath, and UNI, we extract histology features at measured locations using the corresponding pathology foundation model and train a Random Forest regressor for gene expression prediction. Further analyses with additional prediction heads are provided in Supplementary Material~\ref{app:pathology-feature-sufficiency} to assess the predictive capacity of the frozen pathology features. For methods that require a pathology image encoder, we use UNI as the default encoder unless otherwise specified. In particular, Stem, STFlow, HiST, GenAR, and Carta all use UNI features at measured locations. This controls for the visual backbone and makes the comparison focus on how each method models spatial expression from the same histology representation. Additional preprocessing details, baseline settings, and training configurations are provided in Supplementary Material~\ref{app:training-details}.

\FloatBarrier
\subsection{Gene Expression Prediction}

\myparagraph{Overall prediction performance.} 
Tables~\ref{tab:hest-organ-wise}
and~\ref{tab:stimage-organ-wise} report task-level PCC on HEST and STImage-1K4M.   Carta attains the highest average PCC on both resources, reaching $0.425$ on HEST and $0.653$ on STImage-1K4M.   Compared with the strongest baseline, STFlow, Carta improves average PCC by $0.024$ and $0.021$, respectively. The gains are broad across tasks: Carta ranks first on nine of the ten HEST cancer-specific tasks and five of the six STImage-1K4M organ-level tasks, ranking second on IDC and Heart, respectively.

These results demonstrate the efficacy of field reconstruction. 
Because Carta and STFlow use the same UNI histology features~\citep{chen2024towards}, their comparison isolates the expression model rather than the visual backbone. 
Stem predicts expression from local histology patches, while STFlow adds slide-level modeling through local spatial attention and flow matching. 
Carta changes how expression is predicted: regional expression states are estimated first and then used to condition local predictions at the measured coordinates. 
The improvement over STFlow therefore reflects the value of regional-to-local field reconstruction, rather than a stronger image encoder.

\FloatBarrier
\Needspace{0.34\textheight}
\begin{wrapfigure}{r}{0.55\textwidth}
    \vspace{-8pt}
    \centering
    \includegraphics[width=0.52\textwidth]{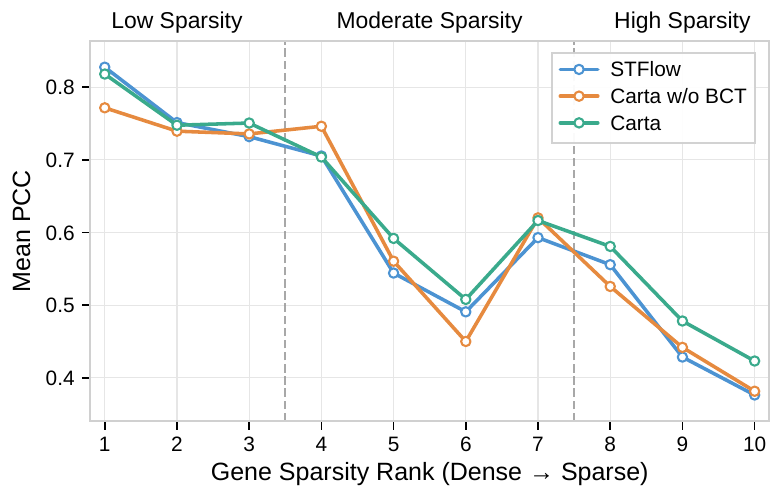}
    \vspace{-5pt}
    \caption{
    \textbf{Gene sparsity-wise PCC on HEST-LUNG.}
    Genes are ranked by zero-expression fraction and grouped into ten bins from dense to sparse. Curves show the mean PCC within each bin.
    }
    \label{fig:sparsity-ranked-pcc}
    \vspace{-5pt}
\end{wrapfigure}

\myparagraph{Gene sparsity-wise comparison.}
\label{app:Gene_sparsity}
Figure~\ref{fig:sparsity-ranked-pcc} presents gene-wise performance across ten sparsity-ranked bins on HEST-LUNG. We select HEST-LUNG because its selected genes exhibit a clear and broad sparsity spectrum, making the dense-to-sparse comparison more informative. While prediction performance generally decreases with increasing sparsity, Carta maintains higher PCC than competing methods, with more pronounced advantages on sparse genes. Moreover, removing BCT leads to a noticeable performance drop in the medium-to-high sparsity regime, highlighting its effectiveness in recovering sparse gene-expression signals. These results demonstrate the benefits of Carta for sparse gene expression prediction, with BCT further improving its performance.

\par
\WFclear

\definecolor{cartagreen}{HTML}{2E7D55}
\definecolor{hestgreenbg}{HTML}{EAF7EF}
\begin{table}[!htbp]
\centering
\footnotesize
\captionsetup{skip=4pt}
\caption{
\textbf{Biomarker prediction.}
PCC is reported for six representative biomarkers; subscripts denote standard deviation.
}

\label{tab:biomarker}
\setlength{\tabcolsep}{6pt}
\renewcommand{\arraystretch}{1.05}

\begin{adjustbox}{max width=\linewidth}
\begin{tabular}{@{}lcccccc@{}}
\toprule
\textbf{Method} 
& \textbf{UBE2C} 
& \textbf{MET} 
& \textbf{KRTDAP} 
& \textbf{PTPRC} 
& \textbf{GATA3} 
& \textbf{ERBB2} \\
\midrule

BLEEP
& $\res{0.618}{.005}$ 
& $\res{0.769}{.006}$ 
& $\res{0.365}{.001}$ 
& $\res{0.634}{.007}$ 
& $\res{0.789}{.012}$ 
& $\res{0.724}{.009}$ \\

HisToGene
& $\res{0.667}{.002}$ 
& $\res{0.561}{.004}$ 
& $\res{0.343}{.001}$ 
& $\res{0.597}{.005}$ 
& $\res{0.721}{.008}$ 
& $\res{0.664}{.003}$ \\

Stem
& $\res{0.453}{.014}$ 
& $\res{0.765}{.006}$ 
& $\res{0.820}{.008}$ 
& $\res{0.812}{.005}$ 
& $\res{0.786}{.007}$ 
& $\res{0.754}{.005}$ \\

STFlow
& $\second{0.749}{.006}$ 
& $\second{0.816}{.007}$ 
& $\second{0.872}{.016}$ 
& $\second{0.851}{.009}$ 
& $\second{0.837}{.004}$ 
& $\second{0.826}{.007}$ \\

STNet
& $\res{0.674}{.004}$ 
& $\res{0.764}{.001}$ 
& $\res{0.433}{.002}$ 
& $\res{0.824}{.003}$ 
& $\res{0.743}{.002}$ 
& $\res{0.707}{.004}$ \\

UNI
& $\res{0.701}{.001}$ 
& $\res{0.745}{.002}$ 
& $\res{0.810}{.005}$ 
& $\res{0.746}{.001}$ 
& $\res{0.822}{.003}$ 
& $\res{0.793}{.006}$ \\

\cellcolor{hestgreenbg}{\textbf{Carta}}
& \cellcolor{hestgreenbg}$\best{0.760}{.004}$
& \cellcolor{hestgreenbg}$\best{0.839}{.002}$
& \cellcolor{hestgreenbg}$\best{0.897}{.011}$
& \cellcolor{hestgreenbg}$\best{0.866}{.005}$
& \cellcolor{hestgreenbg}$\best{0.853}{.007}$
& \cellcolor{hestgreenbg}$\best{0.835}{.006}$ \\
\bottomrule
\end{tabular}
\end{adjustbox}

\end{table}

\myparagraph{Biomarker prediction.}
We further evaluate Carta on six marker genes: UBE2C~\citep{ma2023ube2c}, MET~\citep{gibney2013c}, KRTDAP~\citep{tsuchida2004characterization,su2021psmc2}, PTPRC~\citep{al2021protein,li2023ptprc}, GATA3~\citep{mehra2005identification}, and ERBB2~\citep{revillion1998erbb2}.   Table~\ref{tab:biomarker} reports PCC for each gene.   Carta obtains the highest PCC on all six markers and reaches an average PCC of $0.842$, compared with $0.825$ for STFlow, the strongest baseline.   Marker genes provide a useful stress test because their expression is often localized to specific tissue regions or cell populations; errors that are diluted in average gene-level metrics can become visible at this scale.

\FloatBarrier
\suppressfloats[t]
\begin{figure}[t]
\centering
\includegraphics[width=1.0\textwidth]{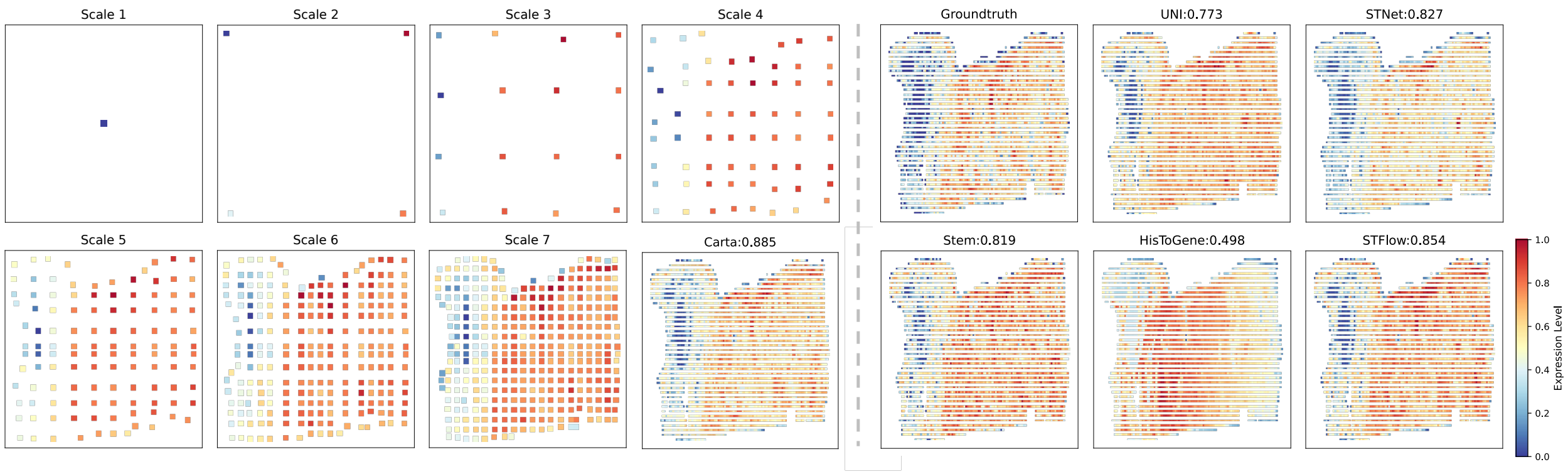}
\vspace{-5pt}
\caption{
\textbf{Biomarker expression maps and scale-wise reconstruction.}
We visualize the coarse-to-fine reconstruction of PTPRC at TENX141 and compare Carta with representative baselines, with PCC reported above the final predictions.
}
\vspace{-5pt}
\label{fig:biomarker}
\end{figure}

\Needspace{0.18\textheight}
Figure~\ref{fig:biomarker} shows representative spatial predictions and the coarse-to-fine reconstruction process.
For PTPRC in TENX141, Carta progressively refines coarse regional expression estimates into a spatially structured expression map. 
Compared with competing methods, Carta preserves the broad high-expression pattern and maintains sharper low-expression boundaries.
These visualizations show that the biomarker gains are not only due to higher global correlation. Carta better preserves where marker expression appears, and where it should remain low, which is consistent with reconstructing expression as a spatial field rather than fitting each measured location independently. Additional biomarker expression maps and coarse-to-fine reconstruction visualizations are provided in Supplementary Material~\ref{app:more_vis_maps} and Supplementary Material~\ref{app:vis-coarse-to-fine}.

\definecolor{defaultorange}{HTML}{FFF1DF}
\definecolor{bestorange}{HTML}{C45A1A}
\definecolor{cartaorange}{HTML}{C87500}  






\FloatBarrier
\Needspace{0.29\textheight}
\subsection{Ablation Studies}
\noindent

\begin{wraptable}{r}{0.45\textwidth}
\begin{minipage}{\linewidth}
\captionsetup{skip=4pt}
    \centering
    \footnotesize
    \vspace{-12pt}
    \caption{\textbf{Component ablation of Carta.}
    We report the average PCC on HEST.}
    
    \label{tab:carta-ablation}
    \setlength{\tabcolsep}{5pt}
    \renewcommand{\arraystretch}{1.10}
    \begin{adjustbox}{max width=\linewidth}
    \begin{tabular}{lccc}
        \toprule
        \textbf{Variant} & \textbf{Ciga} & \textbf{UNI} & \textbf{Gigapath} \\
        \midrule
        w/o MS \& BCT
        & $0.340$ & $0.398$ & $0.386$ \\
        w/o MS
        & $0.346$ & $0.406$ & $0.395$ \\
        w/o BCT
        & $0.351$ & $0.413$ & $0.401$ \\
        \cellcolor{defaultorange}{\textbf{Carta}}
        & \cellcolor{defaultorange}{$\mathbf{0.363}$}
        & \cellcolor{defaultorange}{$\mathbf{0.425}$}
        & \cellcolor{defaultorange}{$\mathbf{0.412}$} \\
        \bottomrule
    \end{tabular}
    \end{adjustbox}
    \par
    
\end{minipage}
\end{wraptable}
\myparagraph{Component ablation.} Table~\ref{tab:carta-ablation} isolates the regional-to-local hierarchy
and BCT regularization under the same training protocol.
With the UNI encoder fixed, adding BCT to the variant without either
component improves the average PCC from $0.398$ to $0.406$, while adding
the multi-scale hierarchy alone yields a larger improvement to $0.413$. 
Combining both components further increases the PCC to $0.425$,
demonstrating that both components contribute and have
complementary effects. 
The same ordering holds with Ciga and Gigapath features, suggesting that
Carta improves expression modeling rather than relying on a specific 
pathology foundation model.

\Needspace{0.25\textheight}
\begin{wraptable}{r}{0.45\textwidth}
\begin{minipage}{\linewidth}
\captionsetup{skip=4pt}
    \centering
    \footnotesize
    \vspace{-12pt}
    \caption{
        \textbf{Cross-scale state propagation on HEST.}
        We compare different propagation strategies and state
        representations. The best results are highlighted in bold,
        and the default setting is shaded.
    }
    
    \label{tab:state-propagation}
    \setlength{\tabcolsep}{4pt}
    \renewcommand{\arraystretch}{1.10}
    \begin{adjustbox}{max width=\linewidth}
    \begin{tabular}{llcc}
        \toprule
        \textbf{State} & \textbf{Transfer}
        & \textbf{PCC} $\uparrow$
        & \textbf{MSE} $\downarrow$ \\
        \midrule
        Expression & Global broadcast
        & $0.414$ & $1.219$ \\
        Expression & Learned attention
        & $0.419$ & $1.208$ \\
        Expression & Nearest parent
        & $0.421$ & $1.206$ \\
        Latent & Distance-weighted
        & $0.419$ & $1.207$ \\

        \rowcolor{hestbg}
        \textbf{Expression}
        & \textbf{Distance-weighted}
        & $\mathbf{0.425}$
        & $\mathbf{1.205}$ \\
        \bottomrule
    \end{tabular}
    \end{adjustbox}
    \par
\end{minipage}
\end{wraptable}
\myparagraph{Cross-scale state propagation.} Table~\ref{tab:state-propagation} compares different cross-scale transfer strategies and state representations under the same training protocol.
Among expression-state variants, distance-weighted propagation achieves the best overall performance, outperforming global broadcast, learned attention, and nearest-parent transfer.
Furthermore, replacing explicit expression states with latent features under the same distance-weighted transfer rule reduces the average PCC from $0.425$ to $0.419$ and increases the average MSE from $1.205$ to $1.207$.
These results demonstrate that both spatially weighted propagation and explicit intermediate expression reconstruction contribute to effective coarse-to-fine field reconstruction.

\FloatBarrier

\section{Conclusion}
We formulated spatial gene expression prediction from histology as field reconstruction, where assayed ST locations are treated as irregular observations of a morphology-conditioned expression field.
The goal is not only to match expression values at measured coordinates, but to preserve the spatial organization of the field they sample.
We presented Carta, which estimates regional expression states from H\&E morphology, refines local expression over measured coordinates, and uses bidirectional conditional transport to shape the distributional structure of the reconstructed field. 
Across 16 organ-specific prediction tasks based on HEST and STImage-1K4M, Carta improves over strong baselines, with gains concentrated on sparse genes and spatially localized expression patterns.
These results suggest that a key challenge is not only missing spatial context, but also supervision limited to sampled values. By treating spatial expression as a field, Carta incorporates regional organization and local heterogeneity into the reconstruction process, leading to more faithful spatial expression reconstruction. Limitations and broader impact are discussed in Supplementary Material~\ref{app:limitations}.

\bibliographystyle{plainnat}
\bibliography{main}





\clearpage



\section*{Supplementary Material}
\appendix

\definecolor{defaultblue}{HTML}{EAF2FF}
\definecolor{bestblue}{HTML}{2F6FC6}

\definecolor{bctbluebg}{HTML}{EEF4FC}

\definecolor{bctorangebg}{HTML}{FFF1DF}

\definecolor{bctgreenbg}{HTML}{EAF7EF}

\definecolor{bctpurplebg}{HTML}{F1ECFA}

\section{Implementation Details}
\label{app:implementation}

This section provides supplementary experimental details for Carta.
We first describe the implementation and training configurations, then introduce the baseline methods used for comparison, and finally present the datasets and evaluation protocols.

\subsection{Training Details and Baseline Settings}
\label{app:training-details}

All experiments follow the preprocessing pipeline described in the main text. 
For each measured spatial location, we crop a $224\times224$ histology patch centered at that location and extract its visual representation using a frozen pathology foundation model.
The resulting patch-level features are used as the input to Carta. 
Unless otherwise specified, we use UNI as the default pathology image encoder. 

For all trainable neural models, we search the learning rate over
$\{10^{-3}, 5\times10^{-4}, 10^{-4}\}$ and the dropout rate over
$\{0, 0.2, 0.5\}$ on \texttt{HEST-1k} under the same cross-validation
protocol. All models are optimized using Adam~\citep{kingma2014adam} for up to 100 epochs, with early stopping using a patience of 20 epochs.  
For \texttt{HEST-1k}, we follow the patient-disjoint cross-validation and
checkpoint-selection protocol of STFlow~\citep{huang2025scalable} under the official HEST-Bench
splits~\citep{jaume2024hest}. 
The same procedure is applied to Carta and all re-trained baselines. 
For \texttt{STImage-1K4M}, the architecture and hyperparameters selected on
\texttt{HEST-1k} are kept fixed without further tuning. All trainable models
are optimized using Adam for a fixed 100 epochs under the same
training budget, and the final-epoch checkpoint is used for
evaluation. 

To stabilize training, gradient norms are clipped to $1.0$ for each model. 
All pathology foundation model parameters are frozen during training. 
All experiments are conducted on a single NVIDIA RTX A6000 GPU.  
To reduce randomness, we repeat the main experiments with $3$ random seeds and report the averaged results.

For Carta, we set the batch size to $2$.
Unless otherwise specified, we use the default multi-scale schedule
$\mathcal{S}_{\mathrm{def}}=\{1^2,2^2,4^2,8^2,12^2,16^2,20^2,N\}$,
where $N$ denotes the native spot-level scale of each tissue section.
For a scale denoted by $s^2$, we use a square spatial partition with
$S_x^{(l)}=S_y^{(l)}=s$, and retain only non-empty bins as regional nodes.
The interpolation bandwidth in cross-scale expression-state propagation is set to $\lambda_u=0.2$. 
The model hidden dimension is set to $d=256$, and the prediction-head intermediate dimension is set to $d_f=256$.
The model uses one cross-attention layer for fusing histology features and propagated expression states, followed by $4$ self-attention Transformer blocks.
The number of attention heads is set to $4$ for both cross-attention and self-attention layers.
The number of nearest neighbors is set to $8$ for both attention and cross-scale KNN interpolation.

For the BCT objective, we set the loss weight to $\lambda_{\mathrm{BCT}}=0.001$, the transport temperature to $\tau=1$, the numerical stability constant to $\delta=10^{-6}$, the source-target balancing coefficient to $\alpha=0.5$, and the momentum coefficient of the dynamic prior to $\beta=0.05$.
The dynamic prior is initialized as a uniform distribution over gene anchors and updated online during training.
These values are used as the default BCT configuration unless otherwise specified.

For the Ciga, Gigapath, and UNI feature baselines, we use the official
pretrained weights of the corresponding pathology foundation models and train
a Random Forest regressor for gene expression prediction. To examine whether
their performance is limited by the choice of prediction head, we additionally
evaluate linear regression, ridge regression, and a two-layer MLP in
Supplementary Material~\ref{app:pathology-feature-sufficiency}.
The number of trees is set to $70$.
For the remaining baselines, we follow their official implementations whenever available.
Stem, STFlow, HiST, and GenAR also use UNI features by default. 
The first three take slide-level inputs with batch size $2$; for the last, we adapt its scale hierarchy to the 50-gene setting using $\{1,2,5,10,25,50\}$. 
For DANet and BLEEP, we construct the expression library using all training-set gene expression profiles, and use the linear combination of the top-$50$ retrieved profiles as the prediction.
For STNet, we use ResNet-50~\citep{he2016deep} as the default image encoder and set the batch size to $128$.
For HisToGene, we follow the official design and use its ViT-based~\citep{dosovitskiy2020image} encoder with slide-level inputs and batch size $2$.
For TRIPLEX, FmH2ST, and MERGE, we preserve their original feature extraction pipelines and core spatial architectures.
All baselines are evaluated using the same samples, data partitions, target genes, expression preprocessing, and evaluation metrics.

\begin{table}[htpb]
	\centering
	\caption{
\textbf{Dataset statistics of HEST-Bench.}
For each cancer-specific subset, we report the organ type, ST technology, number of patients and samples, evaluation splits, and average number of measured spots.
}
	\label{tab:hest-statistics}
	\setlength{\tabcolsep}{4.0pt}
	\renewcommand{\arraystretch}{1.15}
	\scriptsize
	
	\begin{tabular}{@{}c|cccccccccc@{}}
		\toprule
		& \textbf{IDC} & \textbf{PRAD} & \textbf{PAAD} & \textbf{SKCM} & \textbf{COAD} 
		& \textbf{READ} & \textbf{CCRCC} & \textbf{HCC} & \textbf{LUNG} & \textbf{LYMPH} \\
		\midrule

		\textbf{Organ}
		& Breast & Prostate & Pancreas & Skin & Colon & Rectum & Kidney & Liver & Lung 
		& \shortstack{Axillary Lymph Nodes} \\
		
		\textbf{Technology}
		& Xenium & Visium & Xenium & Xenium & Visium & Visium & Visium & Visium & Xenium & Visium \\

		\textbf{\#Patients}
		& 4 & 2 & 3 & 2 & 3 & 2 & 24 & 2 & 2 & 4 \\
		
		\textbf{\#Samples}
		& 4 & 23 & 3 & 2 & 6 & 4 & 24 & 2 & 2 & 4 \\

		\textbf{\#Splits}
		& 4 & 2 & 3 & 2 & 2 & 2 & 6 & 2 & 2 & 4 \\
        
		\rowcolor{datasetrowbg}
		\textbf{Avg. spots}
		& 4925 & 2454 & 2780 & 1741 & 5079 & 1909 & 2792 & 1941 & 1944 & 4990 \\
		\bottomrule
	\end{tabular}
\end{table}

\begin{table}[htpb]
	\centering
	\caption{
\textbf{Dataset statistics of our STImage-derived benchmark.}
We summarize the ST technology, number of patients and tissue samples, and average number of measured spots for each organ-level subset.
}
	\label{tab:stimage-statistics}
	\setlength{\tabcolsep}{4.0pt}
	\renewcommand{\arraystretch}{1.15}
	\footnotesize
	\begin{tabular}{@{}c|cccccc@{}}
		\toprule
		& \textbf{Brain} & \textbf{Breast} & \textbf{Heart} & \textbf{Liver} & \textbf{Pancreas} & \textbf{Skin} \\
		\midrule
		
		\textbf{Technology}
		& Visium & Visium / ST & Visium / ST & Visium 
		& Visium & Visium  \\

        \textbf{\#Patients}
		& 42 & 84 & 29 & 36 & 31 & 26  \\
        
		\textbf{\#Samples}
		& 90 & 195 & 50 & 44 
		& 40 & 49 \\
		
		\rowcolor{datasetrowbg}
		\textbf{Avg. spots}
		& 3259 & 1073 & 2171  
		& 2319 & 1635 & 1235 \\
		\bottomrule
	\end{tabular}
\end{table}
\subsection{Dataset}
\label{app:dataset}

Table~\ref{tab:hest-statistics} presents the dataset statistics of HEST-Bench.
We follow the HEST-Bench~\citep{jaume2024hest} setting used in prior work, which is constructed from HEST-1k and contains ten organ-specific subsets.
HEST-Bench covers diverse tissue types and spatial transcriptomics technologies, including Visium and Xenium.
Following the standard evaluation protocol, we adopt patient-stratified cross-validation on HEST-Bench to avoid data leakage caused by samples from the same patient appearing in both training and test sets.
For each dataset, the model predicts the top 50 highly variable genes, and Pearson correlation coefficient is used as the primary evaluation metric.

In addition to HEST-Bench, we construct an STImage-derived benchmark based on STImage-1K4M~\citep{chen2024stimage}. 
Specifically, we select samples for each organ and use a
patient-disjoint 80/20 train/test split. 
Different from the strictly single-platform setting used in prior work, our constructed benchmark contains organ-specific subsets with both single-platform and partially mixed-platform compositions.
Most organ subsets are collected from a single spatial transcriptomics technology, while some subsets contain samples from both ST and Visium platforms.
This setting introduces moderate platform heterogeneity in part of the benchmark and provides a more challenging evaluation scenario for testing model robustness across tissue and measurement conditions.

To ensure fair comparison, all samples are processed with the same preprocessing pipeline, including image patch extraction, coordinate normalization, gene expression normalization, gene selection, and train/test splitting.
The platform information is not used as an additional supervision signal, but is retained as part of the benchmark difficulty.
For transparency, we report the technology composition of each organ subset in Table~\ref{tab:stimage-statistics}.
As in HEST-Bench, the model predicts the top 50 highly variable genes, and Pearson correlation coefficient is used as the evaluation metric.

\section{Supplementary Derivations of the Method}
\label{app:method-derivations}

This section gives implementation details for Carta that are omitted from the main text. 
We describe coordinate normalization and regional node construction, distance-weighted propagation across scales, the local geometry-aware attention operator, and the gene-anchor prior used in the bidirectional conditional transport objective.


\subsection{Coordinate Binning and Regional Node Aggregation}
\label{app:coordinate-binning}
For each tissue section, Carta constructs regional nodes by grouping measured locations at different
spatial scales. Let the raw coordinate of spot $i$ be
$\mathbf{c}_i^{\mathrm{raw}}=(a_i,b_i)$. We first normalize coordinates
within each tissue section. Let
\begin{equation}
s=\max\left(a_{\max}-a_{\min},\,b_{\max}-b_{\min}\right).
\end{equation}
Then,
\begin{equation}
\tilde a_i
=
\frac{a_i-a_{\min}}{s+\epsilon},
\qquad
\tilde b_i
=
\frac{b_i-b_{\min}}{s+\epsilon}.
\end{equation}
where $\epsilon=10^{-6}$ is used for numerical stability.
We write $\tilde{\mathbf c}_i=(\tilde a_i,\tilde b_i)$.
All binning and distance computations below use the normalized coordinates. At scale $l$, the normalized coordinate frame is divided into
$S_x^{(l)}\times S_y^{(l)}$ bins. 
The bin indices for spot $i$ are:
\begin{equation}
    b_{i,x}^{(l)}
    =
    \min\!\left(\left\lfloor S_x^{(l)}\tilde a_i\right\rfloor, S_x^{(l)}-1\right),
    \qquad
    b_{i,y}^{(l)}
    =
    \min\!\left(\left\lfloor S_y^{(l)}\tilde b_i\right\rfloor, S_y^{(l)}-1\right).
\end{equation}
Using zero-based indexing, the cell index is:
\begin{equation}
    \mathrm{cell}^{(l)}(i)
    =
    b_{i,y}^{(l)}S_x^{(l)} + b_{i,x}^{(l)} .
\end{equation}

We keep cells that contain at least one measured location. 
The $j$-th retained cell at scale $l$ defines a regional node:
$
R_j^{(l)}
=
\{i:\mathrm{cell}^{(l)}(i)=j\}$.
For each regional node, we average the histology embeddings and normalized coordinates of the locations it contains:
\begin{equation}
    h_j^{(l)}
    =
    \frac{1}{|R_j^{(l)}|}
    \sum_{i\in R_j^{(l)}}h_i,
    \qquad
    \tilde c_j^{(l)}
    =
    \frac{1}{|R_j^{(l)}|}
    \sum_{i\in R_j^{(l)}}\tilde c_i .
\end{equation}
For the finest scale, each node corresponds to one measured ST location, so $h_i^{(L)}=h_i, \, \tilde c_i^{(L)}=\tilde c_i$.
This construction keeps the hierarchy tied to measured assay coordinates rather than to a regular image grid.

\subsection{Cross-Scale Expression-State Propagation}
\label{app:cross-scale}

For $l\geq 2$, expression states are passed from scale $l-1$ to scale $l$ by distance-weighted interpolation. 
For node $i$ at scale $l$, let $\mathcal{N}_i^{(l)}$ denote its $K$ nearest nodes from scale $l-1$, using normalized coordinates. 
The propagated expression state is:
\begin{equation}
    p_i^{(l)}
    =
    \sum_{j\in\mathcal{N}_i^{(l)}}
    w_{ij}^{(l)} \hat y_j^{(l-1)},
\end{equation}
where:
\begin{equation}
    w_{ij}^{(l)}
    =
    \frac{
    \exp\!\left(-\|\tilde c_i^{(l)}-\tilde c_j^{(l-1)}\|_2^2/\lambda_u\right)
    }{
    \sum_{j'\in\mathcal{N}_i^{(l)}}
    \exp\!\left(-\|\tilde c_i^{(l)}-\tilde c_{j'}^{(l-1)}\|_2^2/\lambda_u\right)
    } .
\end{equation}
The bandwidth $\lambda_u>0$ controls the locality of interpolation. 
Smaller values emphasize nearby nodes, while larger values average over a wider neighborhood.

At the coarsest scale, no previous scale is available. 
As described in the main text, Carta initializes the expression condition with a learnable gene-wise Gaussian prior:
\begin{equation}
    p^{(1)}
    =
    \mu+\exp(\log\sigma)\odot\epsilon,
    \qquad
    \epsilon\sim\mathcal{N}(0,I).
\end{equation}
For every coarsest-scale node $i$, we set $p_i^{(1)}=p^{(1)}$ during training. 
At evaluation time, we use the mean state $p_i^{(1)}=\mu$ unless otherwise specified.

\subsection{Geometry-Aware Attention Module}
\label{app:geo-attention}

Carta uses geometry-aware attention to exchange information among nearby nodes. 
The operator~\citep{huang2025scalable} is used in two places in Carta: cross-scale conditional fusion and within-scale self-attention. 
The two modules share the same geometric attention form but differ in how queries, keys, and values are constructed.

Let $\mathcal{M}_i^{(l)}$ denote the within-scale spatial neighborhood of node $i$ at scale $l$.
For a neighbor $j\in\mathcal{M}_i^{(l)}$, we first compute the relative coordinate and distance:
\begin{equation}
    r_{ij}^{(l)}=\tilde c_j^{(l)}-\tilde c_i^{(l)},
    \qquad
    d_{ij}^{(l)}=\|r_{ij}^{(l)}\|_2 .
\end{equation}
We then transform $r_{ij}^{(l)}$ into a set of frame-aware coordinate features
$\{\phi_m(r_{ij}^{(l)})\}_{m=1}^{M}$.
The geometric edge feature is then computed as
\begin{equation}
    \rho_{ij}^{(l)}
    =
    \frac{1}{M}
    \sum_{m=1}^{M}
    \psi_e\left(
    \left[
    \phi_m(r_{ij}^{(l)}), d_{ij}^{(l)}
    \right]
    \right),
\end{equation}
where $\psi_e(\cdot)$ is an MLP edge encoder and $M$ is the number of frames.
We use $\rho_{ij}^{(l)}$ for geometric edge features to avoid confusion with the expression-condition embedding defined below. 

For cross-scale conditional fusion, the histology representation provides the query, and the propagated expression state provides keys and values. 
Since $p_i^{(l)}\in\mathbb{R}^{G}$ and the attention module operates in the model dimension, we first project the expression condition:
\begin{equation}
    e_i^{(l)} = W_p p_i^{(l)} \in \mathbb{R}^{d}.
\end{equation}
The query, key, and value are then:
\begin{equation}
    q_i^{\mathrm{cross}}
    =
    W_q^{\mathrm{cross}}\mathrm{LN}(z_i^{(l)}),
    \quad
    [k_j^{\mathrm{cross}},v_j^{\mathrm{cross}}]
    =
    W_{kv}^{\mathrm{cross}}\mathrm{LN}(e_j^{(l)}),
    \quad j\in\mathcal{M}_i^{(l)} .
\end{equation}

For within-scale self-attention, query, key, and value are all generated from the current node features:
\begin{equation}
    [q_i^{\mathrm{self}},k_i^{\mathrm{self}},v_i^{\mathrm{self}}]
    =
    W_{qkv}^{\mathrm{self}}\mathrm{LN}(z_i^{(l)}).
\end{equation}

Given a query $q_i$, neighbor keys $\{k_j\}$, values $\{v_j\}$, and edge features $\{\rho_{ij}^{(l)}\}$, the geometry-aware attention logit is:
\begin{equation}
    a_{ij}
    =
    \psi_a([q_i,k_j,\rho_{ij}^{(l)}]),
\end{equation}
where $\psi_a(\cdot)$ is an MLP that outputs a scalar logit. 
The attention weights are normalized within the neighborhood:
\begin{equation}
    A_{ij}
    =
    \frac{\exp(a_{ij})}
    {\sum_{j'\in\mathcal{M}_i^{(l)}}\exp(a_{ij'})}.
\end{equation}
The message aggregates both value features and geometric edge features:
\begin{equation}
    m_i
    =
    \psi_o
    \left(
    \left[
    \sum_{j\in\mathcal{M}_i^{(l)}}A_{ij}v_j,
    \sum_{j\in\mathcal{M}_i^{(l)}}A_{ij}\rho_{ij}^{(l)}
    \right]
    \right),
\end{equation}
where $\psi_o(\cdot)$ maps the concatenated vector back to the model dimension.

In cross-scale conditional fusion, this message is denoted by $m_i^{\mathrm{cross}}$ and is used to inject the propagated expression condition into the histology representation:
\begin{equation}
    \bar z_i^{(l)}
    =
    z_i^{(l)} + m_i^{\mathrm{cross}},
    \qquad
    \tilde z_i^{(l)}
    =
    \bar z_i^{(l)}
    +
    \mathrm{FFN}_{\mathrm{cross}}\!\left(\mathrm{LN}(\bar z_i^{(l)})\right).
\end{equation}
Here, $\tilde z_i^{(l)}$ denotes the node feature after conditioning on the transferred expression state.

The conditioned features are then passed to the within-scale encoder. 
Here we set:
\begin{equation}
    z_{i,0}^{(l)}=\tilde z_i^{(l)}.
\end{equation}
For the $r$-th within-scale encoder layer, geometry-aware self-attention produces a message $m_{i,r}^{\mathrm{self}}$, followed by an FFN update:
\begin{equation}
    z_{i,r+\frac{1}{2}}^{(l)}
    =
    z_{i,r}^{(l)} + m_{i,r}^{\mathrm{self}},
    \qquad
    z_{i,r+1}^{(l)}
    =
    z_{i,r+\frac{1}{2}}^{(l)}
    +
    \mathrm{FFN}_{\mathrm{self}}\!\left(\mathrm{LN}(z_{i,r+\frac{1}{2}}^{(l)})\right).
\end{equation}
After $R$ within-scale encoder layers, the output feature is:
\begin{equation}
    u_i^{(l)} = z_{i,R}^{(l)}.
\end{equation}
This feature is passed to the prediction head at scale $l$.

\subsection{Gene Anchors and Dynamic Prior Updates in BCT}
\label{app:bct-anchor-prior}

BCT is applied between the intermediate features of the prediction head and gene anchors defined by the output layer.
At scale $l$, the prediction head first produces an intermediate feature:
\begin{equation}
    f_i^{(l)}
    =
    \phi_1(u_i^{(l)})
    =
    \mathrm{LN}
    \left(
    \mathrm{GELU}
    \left(
    W_1 u_i^{(l)} + b_1
    \right)
    \right)
    \in \mathbb{R}^{d_f}.
\end{equation}
The expression prediction is:
\begin{equation}
    \hat y_i^{(l)}
    =
    \psi\left(W_o f_i^{(l)} + b_o\right),
\end{equation}
where $W_1\in\mathbb{R}^{d_f\times d}$, $b_1\in\mathbb{R}^{d_f}$, $W_o\in\mathbb{R}^{G\times d_f}$, and $b_o\in\mathbb{R}^{G}$.
The activation $\psi$ is set to softplus for non-negative expression targets and to the identity map for standardized targets.

Each row of the shared output projection is treated as a gene anchor:
\begin{equation}
    g_k = \operatorname{sg}\!\left(W_o[k,:]\right), \quad k=1,\ldots,G.
\end{equation}
Thus, the same latent directions used to predict gene expression also define the gene-anchor space in BCT.
Given node feature $f_i^{(l)}$ and gene anchor $g_k$, the similarity matrix is:
\begin{equation}
    S_{ki}^{(l)}
    =
    g_k^\top f_i^{(l)}.
\end{equation}
The corresponding cosine transport cost is:
\begin{equation}
    D_{ki}^{(l)}
    =
    1-
    \frac{g_k^\top f_i^{(l)}}
    {\|g_k\|_2\|f_i^{(l)}\|_2}.
\end{equation}

Because gene activity can be uneven within a mini-batch, BCT maintains a dynamic prior
$\mathbf q\in\Delta^{G-1}$ over gene anchors, where:
\[
    \Delta^{G-1}
    =
    \left\{
    \mathbf q\in\mathbb{R}_+^G:
    \sum_{k=1}^{G}q_k=1
    \right\}.
\]
The prior is initialized uniformly:
\begin{equation}
    q_k^{(0)}=\frac{1}{G},
    \qquad
    k=1,\ldots,G.
\end{equation}

During each forward pass, we compute detached node-to-anchor assignments using the current prior:
\begin{equation}
    \pi_{ki}^{\mathrm{old},(l)}
    =
    \mathrm{softmax}_{k}
    \left(
    \frac{\mathrm{sg}(S_{ki}^{(l)})}{\tau}
    +
    \log(q_k+\delta)
    \right),
\end{equation}
where $\tau$ is the temperature, and $\delta=10^{-6}$ is used for numerical stability.

We estimate the batch-level gene-anchor usage by averaging these detached assignments over the nodes used in the current BCT computation at scale $l$:
\begin{equation}
    \bar q_k^{(l)}
    =
    \frac{1}{N^{(l)}}
    \sum_{i=1}^{N^{(l)}}
    \pi_{ki}^{\mathrm{old},(l)}.
\end{equation}
The prior is updated by momentum outside backpropagation:
\begin{equation}
    q_k
    \leftarrow
    (1-\beta)q_k+\beta\bar q_k^{(l)}.
\end{equation}
Since both $\mathbf q$ and $\bar{\mathbf q}^{(l)}$ lie in the probability simplex, this update preserves normalization up to numerical precision.
In implementation, we renormalize $\mathbf q$ after the update.
The updated prior is used in the node-to-anchor transport weights:
\begin{equation}
    \pi_{ki}^{s,(l)}
    =
    \mathrm{softmax}_{k}
    \left(
    \frac{S_{ki}^{(l)}}{\tau}
    +
    \log(q_k+\delta)
    \right).
\end{equation}
The corresponding loss is:
\begin{equation}
    \mathcal{L}_{s}^{(l)}
    =
    \frac{\alpha}{N^{(l)}}
    \sum_{i=1}^{N^{(l)}}\sum_{k=1}^{G}
    D_{ki}^{(l)}\pi_{ki}^{s,(l)}.
\end{equation}
This term assigns each spatial node to gene anchors while accounting for gene-level imbalance through the dynamic prior.
The reverse transport weights assign each gene anchor to spatial nodes:
\begin{equation}
    \pi_{ki}^{t,(l)}
    =
    \mathrm{softmax}_{i}
    \left(
    \frac{S_{ki}^{(l)}}{\tau}
    \right).
\end{equation}
The reverse-direction loss is:
\begin{equation}
    \mathcal{L}_{t}^{(l)}
    =
    (1-\alpha)
    \sum_{k=1}^{G}q_k
    \sum_{i=1}^{N^{(l)}}
    D_{ki}^{(l)}\pi_{ki}^{t,(l)}.
\end{equation}
This direction encourages each gene anchor to be supported by spatial nodes, rather than concentrating the alignment on a small set of frequent responses. 
The parameter $\alpha\in[0,1]$ balances the two transport directions.
The BCT loss at scale $l$ is:
\begin{equation}
    \mathcal{L}_{\mathrm{BCT}}^{(l)}
    =
    \mathcal{L}_{s}^{(l)}
    +
    \mathcal{L}_{t}^{(l)}.
\end{equation}
The multi-scale BCT regularizer is:
\begin{equation}
    \mathcal{L}_{\mathrm{BCT}}
    =
    \frac{1}{L}
    \sum_{l=1}^{L}
    \mathcal{L}_{\mathrm{BCT}}^{(l)}.
\end{equation}
Regression supervision is applied at the finest scale:
\begin{equation}
    \mathcal{L}_{\mathrm{reg}}
    =
    \mathrm{MSE}(\hat{\mathbf Y}^{(L)},\mathbf Y).
\end{equation}
The final objective is:
\begin{equation}
    \mathcal{L}
    =
    \mathcal{L}_{\mathrm{reg}}
    +
    \lambda_{\mathrm{BCT}}\mathcal{L}_{\mathrm{BCT}},
\end{equation}
where $\lambda_{\mathrm{BCT}}$ controls the strength of transport regularization.

\section{Additional Experiments}
\label{app:additional-experiments}

This section provides additional experiments to further characterize Carta and
its key design choices. We first examine stronger prediction heads built on
frozen pathology features. We then evaluate the field-reconstruction formulation
through prediction at held-out spatial coordinates and spatial field diagnostics.
Next, we examine the regional-to-local reconstruction design, including the scale
schedule, cross-scale fusion, the coarsest-scale initial state, and scale-wise
supervision. We further analyze BCT through hyperparameter sensitivity,
gene-anchor construction, and the relationship between gene-anchor geometry
and gene co-expression. Finally, we report computational efficiency. Together,
these experiments complement the main results and provide a more complete
characterization of Carta's reconstruction behavior, design choices, and
computational cost.

\subsection{Ablation on Pathology Features}
\label{app:pathology-feature-sufficiency}

Table~\ref{tab:pfm-heads} compares different prediction heads applied
to the same frozen pathology features from Ciga, Gigapath, and UNI. We evaluate
Random Forest, linear regression, ridge regression, and a two-layer MLP while
keeping the pretrained feature extractors fixed. Hyperparameters are selected on HEST following the same
cross-validation and model-selection protocol as in the main experiments.

\begin{table}[htbp]
\centering
\caption{
\textbf{Comparison of prediction heads on frozen pathology
foundation-model features.}
Average PCC on HEST is reported for Ciga, Gigapath, and UNI.
}
\label{tab:pfm-heads}
\footnotesize
\setlength{\tabcolsep}{7pt}
\renewcommand{\arraystretch}{1.15}
\begin{tabular}{l|ccc}
\toprule
\textbf{Predictor} & \textbf{Ciga} & \textbf{Gigapath} & \textbf{UNI} \\
\midrule
Random Forest     & 0.281 & 0.327 & 0.332 \\
Linear regression & 0.249 & 0.232 & 0.269 \\
Ridge regression  & 0.294 & 0.346 & 0.342 \\
2-layer MLP       & 0.303 & 0.353 & 0.356 \\

\cellcolor{hestbg}{Carta}             & \cellcolor{hestbg}{\textbf{0.363}} & \cellcolor{hestbg}{\textbf{0.412}} & \cellcolor{hestbg}{\textbf{0.425}} \\
\bottomrule
\end{tabular}
\end{table}

Across all three pathology encoders, the feature-only predictors remain below
Carta. The best average PCC achieved using frozen Ciga, Gigapath, and UNI
features is $0.303$, $0.353$, and $0.356$, respectively, compared with
$0.363$, $0.412$, and $0.425$ for Carta. This corresponds to gains of
$0.060$, $0.059$, and $0.069$ PCC. These results indicate that pathology
features alone do not fully explain the prediction performance, and that the
spatial expression modeling in Carta provides additional predictive information.

\begin{table*}[t]
\centering
\caption{
\textbf{Prediction under partial spatial support.}
We evaluate prediction at held-out coordinates under random and
contiguous spatial masking. All methods within each setting are
evaluated on the same held-out coordinates, with results averaged
over five fixed masking seeds.
}
\label{tab:heldout-coordinates}
\small
\setlength{\tabcolsep}{18pt}
\renewcommand{\arraystretch}{1.12}
\begin{tabular}{l|ccc|ccc}
\toprule
\textbf{Method}
& \multicolumn{3}{c|}{\textbf{Random masking}}
& \multicolumn{3}{c}{\textbf{Contiguous masking}} \\
& \textbf{$10\%$}
& \textbf{$25\%$}
& \textbf{$50\%$}
& \textbf{$10\%$}
& \textbf{$25\%$}
& \textbf{$50\%$} \\
\midrule

UNI+RF
& 0.3320 & 0.3312 & 0.3316
& 0.2516 & 0.2771 & 0.2970 \\

kNN-Pred
& 0.3373 & 0.3320 & 0.3205
& 0.1378 & 0.1313 & 0.1220 \\

STFlow
& \underline{0.3838}
& \underline{0.3834}
& \underline{0.3836}
& \underline{0.3033}
& \underline{0.3232}
& \underline{0.3459} \\

\rowcolor{hestbg}
\textbf{Carta}
& \textbf{0.4101}
& \textbf{0.4093}
& \textbf{0.4080}
& \textbf{0.3267}
& \textbf{0.3510}
& \textbf{0.3743} \\
\bottomrule
\end{tabular}
\end{table*}

\subsection{Field Reconstruction from Partial Spatial Support}
\label{app:heldout-coordinates}

To evaluate Carta under incomplete spatial support, we hold out
10\%, 25\%, or 50\% of measured coordinates using either random
or contiguous spatial masking. For contiguous masking, we randomly select a spot as the center
and mask its nearest neighbors by Euclidean distance until the
specified ratio is reached. These locations are excluded from
the finest-scale support, coarse-scale binning, and attention
neighborhoods, and their expression labels are used only for evaluation.
At inference, each held-out location is predicted from its H\&E patch
together with the expression context reconstructed from the retained
coordinates. This setting tests whether Carta can recover omitted
spatial expression under both randomly distributed and spatially
concentrated missing support.

As shown in Table~\ref{tab:heldout-coordinates}, Carta achieves the
highest PCC across all masking ratios under both settings, outperforming
STFlow by $0.0255$ and $0.0265$ PCC on average under random and
contiguous masking, respectively.
Under random masking, Carta maintains stable performance as the
held-out ratio increases from $10\%$ to $50\%$.
Under contiguous masking, Carta consistently outperforms the compared
methods even when entire spatial regions are excluded.
UNI+RF, which predicts expression from local histology alone, and
kNN-Pred, which interpolates from retained spatial locations, remain
substantially weaker. 
These results demonstrate the effectiveness of Carta in reconstructing spatial gene expression fields under incomplete spatial support, even when entire spatial regions are missing.

\begin{table}[htpb]
\centering
\footnotesize
\caption{
\textbf{Spatial field diagnostics on HEST.}
We report regional error, variance consistency, and recovery of spatially selective activation.
}
\label{tab:field-level-diagnostics}
\setlength{\tabcolsep}{5.2pt}
\renewcommand{\arraystretch}{1.12}
\begin{tabular}{l|cccc}
\toprule
\textbf{Method}
& \textbf{Region MAE} $\downarrow$
& \textbf{Var. Log Error} $\downarrow$
& \textbf{Top-10 IoU} $\uparrow$
& \textbf{Top-10 Recall} $\uparrow$ \\
\midrule
STFlow
& 0.524 & 2.965 & 0.176 & 0.284 \\
Carta w/o BCT
& 0.513 & 2.823 & 0.180 & 0.282 \\
\rowcolor{bctbluebg}
\textbf{Carta}
& \textbf{0.487}
& \textbf{2.651}
& \textbf{0.191}
& \textbf{0.293} \\
\bottomrule
\end{tabular}
\vspace{-4pt}
\end{table}

\subsection{Spatial Field Reconstruction Diagnostics}
\label{app:field-level-analysis}
\paragraph{Diagnostics beyond gene-wise correlation.}
Since PCC mainly measures gene-wise correlation, we further introduce several spatial field diagnostics to evaluate whether the reconstructed expression maps preserve regional structure and localized activation patterns.
\textit{Region MAE} first partitions each tissue section into spatial regions using K-means on normalized coordinates and compares the average predicted and measured expression within each region; it measures whether the regional expression intensity is correctly reconstructed.
\textit{Var. Log Error} compares the variance of predicted and measured regional expression in log space; a lower value indicates that the model better preserves field-level variability and is less prone to over-smoothing or noisy over-dispersion.
\textit{Top-10 IoU} computes the intersection-over-union between the predicted and ground-truth top-$10\%$ high-expression locations, measuring the spatial overlap of localized activation regions.
\textit{Top-10 Recall} measures the fraction of ground-truth top-$10\%$ high-expression locations covered by the predicted top-$10\%$ regions, reflecting whether the model misses true localized activation.
All metrics are first computed within each test slide and then aggregated at the organ level.

\paragraph{Results on field-level structure.}
Table~\ref{tab:field-level-diagnostics} shows that Carta improves over STFlow on all four spatial field diagnostics.
Carta reduces Region MAE from $0.524$ to $0.487$, indicating better reconstruction of regional expression intensity.
It also lowers Var. Log Error from $2.965$ to $2.651$, suggesting that the predicted field better preserves expression variability rather than collapsing to an over-smoothed map.
For spatially selective activation, Carta improves both Top-10 IoU and Top-10 Recall, showing better recovery of localized activated regions.
Carta w/o BCT consistently lies between STFlow and the full model on most metrics, indicating that BCT further improves the preservation of regional expression structure and localized activation patterns. 
These results support our formulation of spatial gene expression prediction as field reconstruction rather than independent coordinate-level regression.

\begin{table*}[htpb]
\centering
\tiny
\caption{
\textbf{Ablation study on multi-scale schedule design.}
We report organ-wise PCC on HEST under different multi-scale schedules.
$N$ denotes the native spot-level scale of each tissue section.
}
\label{tab:scale-ablation}
\setlength{\tabcolsep}{3.4pt}
\renewcommand{\arraystretch}{1.18}

\resizebox{\textwidth}{!}{
\begin{tabular}{@{}l|c|ccccccccccc@{}}
\toprule
\textbf{Strategy} & \textbf{Schedule}
& \textbf{CCRCC} & \textbf{COAD} & \textbf{HCC} & \textbf{IDC} & \textbf{LUNG}
& \textbf{LYMPH} & \textbf{PAAD} & \textbf{PRAD} & \textbf{READ} & \textbf{SKCM}
& \textbf{Avg.} \\
\midrule

Very sparse
& $\mathcal{S}_{\mathrm{sp}}$
& $0.328$ & $0.347$ & $0.120$ & $0.581$ & $0.613$
& $0.303$ & $0.503$ & $0.406$ & $0.243$ & $0.691$ & $0.414$ \\

w/o early coarse
& $\mathcal{S}_{\mathrm{wc}}$
& $0.327$ & $0.339$ & $0.118$ & $0.577$ & $0.610$
& $0.308$ & $0.507$ & $0.399$ & $0.237$ & $0.685$ & $0.411$ \\

Finer late
& $\mathcal{S}_{\mathrm{fine}}$
& $\underline{0.339}$ & $\mathbf{0.369}$ & $\underline{0.130}$ & $0.594$ & $\mathbf{0.624}$
& $\mathbf{0.318}$ & $\mathbf{0.518}$ & $\underline{0.420}$ & $\underline{0.258}$ & $\mathbf{0.710}$ & $\mathbf{0.428}$ \\

Larger pre-native
& $\mathcal{S}_{\mathrm{large}}$
& $\mathbf{0.340}$ & $0.357$ & $0.128$ & $\mathbf{0.596}$ & $\underline{0.622}$
& $0.310$ & $0.514$ & $0.419$ & $\mathbf{0.260}$ & $0.695$ & $0.424$ \\

\cellcolor{hestbg}{\textbf{Default}}
& \cellcolor{hestbg}$\mathbf{\mathcal{S}_{\mathrm{def}}}$
& \cellcolor{hestbg}$0.335$
& \cellcolor{hestbg}$\underline{0.363}$
& \cellcolor{hestbg}$\mathbf{0.132}$
& \cellcolor{hestbg}$\underline{0.595}$
& \cellcolor{hestbg}$0.621$
& \cellcolor{hestbg}$\underline{0.315}$
& \cellcolor{hestbg}$\underline{0.516}$
& \cellcolor{hestbg}$\mathbf{0.422}$
& \cellcolor{hestbg}$0.245$
& \cellcolor{hestbg}$\underline{0.703}$
& \cellcolor{hestbg}$\underline{0.425}$ \\
\bottomrule
\end{tabular}
}

\vspace{2pt}
\begin{minipage}{0.98\textwidth}
\scriptsize

\textit{Schedule definitions.}
$\mathcal{S}_{\mathrm{sp}}=\{1^2,8^2,20^2,N\}$;
$\mathcal{S}_{\mathrm{wc}}=\{8^2,12^2,16^2,20^2,N\}$;
$\mathcal{S}_{\mathrm{def}}=\{1^2,2^2,4^2,8^2,12^2,16^2,20^2,N\}$;
$\mathcal{S}_{\mathrm{fine}}=\{1^2,2^2,4^2,8^2,10^2,12^2,14^2,16^2,18^2,20^2,N\}$;
$\mathcal{S}_{\mathrm{large}}=\{1^2,2^2,4^2,8^2,12^2,16^2,20^2,28^2,N\}$.
\end{minipage}
\end{table*}

\subsection{Scale Schedule Design}
\label{app:scale-schedule}
Table~\ref{tab:scale-ablation} studies how the choice of scale schedule affects field reconstruction.   The two reduced schedules perform worse: the sparse schedule $\mathcal{S}_{\mathrm{sp}}$ reaches an average PCC of $0.414$, and the schedule without early coarse stages $\mathcal{S}_{\mathrm{wc}}$ reaches $0.411$.   These drops indicate that Carta benefits from both early regional states and intermediate transitions between regional and local predictions.
Adding denser late-stage scales improves the average PCC to $0.428$ for $\mathcal{S}_{\mathrm{fine}}$, showing that refinement near the measured coordinates is useful.   The larger pre-final schedule $\mathcal{S}_{\mathrm{large}}$ reaches $0.424$, suggesting that adding a single larger intermediate layout is less effective than maintaining a balanced sequence of regional and near-coordinate scales.   The default schedule $\mathcal{S}_{\mathrm{def}}$ obtains $0.425$ average PCC, within $0.003$ of $\mathcal{S}_{\mathrm{fine}}$, while using a more compact scale sequence.   We therefore use $\mathcal{S}_{\mathrm{def}}$ in the main experiments.

\begin{table*}[htbp]
	\centering
	\tiny
	\caption{
		\textbf{Organ-wise ablation study of cross-scale fusion strategies on HEST.}
		We report PCC and MSE for each organ using UNI as the default pathology foundation model.
		All variants use the same training configuration, multi-scale reconstruction schedule, and BCT configuration, except for the way propagated expression states are fused with current-scale histology features.
		Best and second-best results are highlighted in bold and underlined, respectively.
	}
	\label{tab:cross-fusion-strategy-ablation}
	
	\setlength{\tabcolsep}{3.9pt}
	\renewcommand{\arraystretch}{1.14}
	
	\resizebox{\textwidth}{!}{
		\begin{tabular}{@{}ll*{11}{c}@{}}
			\toprule
			\multirow{2}{*}{\textbf{Metric}} 
			& \multirow{2}{*}{\textbf{Fusion strategy}}
			& \multicolumn{10}{c}{\textbf{Organ}} 
			& \multirow{2}{*}{\textbf{Avg.}} \\
			\cmidrule(lr){3-12}
			& 
			& \textbf{CCRCC} & \textbf{COAD} & \textbf{HCC} & \textbf{IDC} & \textbf{LUNG}
			& \textbf{LYMPH} & \textbf{PAAD} & \textbf{PRAD} & \textbf{READ} & \textbf{SKCM}
			& \\
			\midrule
			
			\multirow{3}{*}{
				\begin{tabular}[c]{@{}c@{}}
					\textbf{PCC}\\[-0.15em]
					$\uparrow$
				\end{tabular}
			}
			& Concat + Linear
			& $\underline{0.315}$ & $\mathbf{0.364}$ & $0.119$ & $\underline{0.591}$ & $\underline{0.620}$
			& $\underline{0.310}$ & $\underline{0.506}$ & $0.418$ & $\underline{0.239}$ & $0.695$
			& $\underline{0.418}$ \\
			
			& Linear + Add
			& $0.309$ & $0.360$ & $\underline{0.121}$ & $0.590$ & $0.618$
			& $0.306$ & $0.505$ & $\underline{0.420}$ & $0.234$ & $\underline{0.700}$
			& $0.416$ \\
			
			& \cellcolor{hestbg}\textbf{Cross-attention}
			& \cellcolor{hestbg}$\mathbf{0.335}$
			& \cellcolor{hestbg}$\underline{0.363}$
			& \cellcolor{hestbg}$\mathbf{0.132}$
			& \cellcolor{hestbg}$\mathbf{0.595}$
			& \cellcolor{hestbg}$\mathbf{0.621}$
			& \cellcolor{hestbg}$\mathbf{0.315}$
			& \cellcolor{hestbg}$\mathbf{0.516}$
			& \cellcolor{hestbg}$\mathbf{0.422}$
			& \cellcolor{hestbg}$\mathbf{0.245}$
			& \cellcolor{hestbg}$\mathbf{0.703}$
			& \cellcolor{hestbg}$\mathbf{0.425}$ \\
			
			\midrule
			
			\multirow{3}{*}{
				\begin{tabular}[c]{@{}c@{}}
					\textbf{MSE}\\[-0.15em]
					$\downarrow$
				\end{tabular}
			}
			& Concat + Linear
			& $0.516$ & $\mathbf{1.828}$ & $\mathbf{0.403}$ & $2.689$ & $\mathbf{2.016}$
			& $\mathbf{0.492}$ & $1.382$ & $\mathbf{0.538}$ & $\mathbf{0.301}$ & $1.925$
			& $\underline{1.209}$ \\
			
			& Linear + Add
			& $\underline{0.514}$ & $\underline{1.837}$ & $\underline{0.404}$ & $\underline{2.676}$ & $\mathbf{2.016}$
			& $0.542$ & $\underline{1.380}$ & $\underline{0.540}$ & $\underline{0.304}$ & $\underline{1.906}$
			& $1.212$ \\
			
			& \cellcolor{hestbg}\textbf{Cross-attention}
			& \cellcolor{hestbg}$\mathbf{0.508}$
			& \cellcolor{hestbg}$1.853$
			& \cellcolor{hestbg}$0.415$
			& \cellcolor{hestbg}$\mathbf{2.656}$
			& \cellcolor{hestbg}$\underline{2.018}$
			& \cellcolor{hestbg}$\underline{0.528}$
			& \cellcolor{hestbg}$\mathbf{1.346}$
			& \cellcolor{hestbg}$0.550$
			& \cellcolor{hestbg}$0.305$
			& \cellcolor{hestbg}$\mathbf{1.872}$
			& \cellcolor{hestbg}$\mathbf{1.205}$ \\
			
			\bottomrule
		\end{tabular}
	}
\end{table*}

\subsection{Cross-scale Fusion Strategy}
\label{app:cross-fusion-strategy-ablation}
Table~\ref{tab:cross-fusion-strategy-ablation} compares different strategies for fusing propagated coarse-scale expression states with current-scale histology features.
The concat + linear strategy directly concatenates the two representations and projects them back to the hidden space, while the linear + add strategy first maps the expression state into the histology feature space and then performs residual addition.
Although these two simple fusion strategies can inject coarse-scale expression information, they treat the propagated expression state as a fixed feature-level condition and do not explicitly model local spatial correspondence during fusion.

In contrast, the default cross-attention fusion uses current-scale histology features as queries and propagated expression states as keys and values, with geometric edge features incorporated into the attention computation.
This allows each node to selectively aggregate expression conditions from its local spatial neighborhood in a content-adaptive and geometry-aware manner.
As shown in Table~\ref{tab:cross-fusion-strategy-ablation}, cross-attention achieves the best average PCC of $0.425$, outperforming concat + linear fusion ($0.418$) and linear + add fusion ($0.416$).
It also obtains the lowest average MSE of $1.205$.

The organ-wise results further show that cross-attention brings consistent gains in PCC across most organs, achieving the best result on nine out of ten organs.
This suggests that adaptive cross-scale conditioning is more effective for preserving relative spatial expression variation than simple feature concatenation or addition.
Meanwhile, concat + linear and linear + add obtain competitive MSE on some organs, indicating that simple fusion may still reduce point-wise errors in certain cases.
However, their lower PCC suggests weaker recovery of spatial expression patterns.
Overall, these results support our design choice of using local geometry-aware cross-attention for cross-scale expression fusion in Carta.

\begin{table*}[htbp]
	\centering
	\tiny
	\caption{
		\textbf{Organ-wise ablation study of the initial expression state on HEST.}
		We report PCC and MSE for each organ using UNI as the default pathology foundation model.
		All variants use the same multi-scale reconstruction schedule and BCT configuration.
		They differ only in whether a small stochastic perturbation is injected into the learnable coarsest-scale expression state during training.
		Evaluation is performed with deterministic initialization.
		Best results are highlighted in bold, and the default setting is shaded.
	}
	\label{tab:initial-prior-ablation}
	
	\setlength{\tabcolsep}{3.9pt}
	\renewcommand{\arraystretch}{1.14}
	
	\resizebox{\textwidth}{!}{
		\begin{tabular}{@{}ll*{11}{c}@{}}
			\toprule
			\multirow{2}{*}{\textbf{Metric}} 
			& \multirow{2}{*}{\textbf{Initial state during training}}
			& \multicolumn{10}{c}{\textbf{Organ}} 
			& \multirow{2}{*}{\textbf{Avg.}} \\
			\cmidrule(lr){3-12}
			& 
			& \textbf{CCRCC} & \textbf{COAD} & \textbf{HCC} & \textbf{IDC} & \textbf{LUNG}
			& \textbf{LYMPH} & \textbf{PAAD} & \textbf{PRAD} & \textbf{READ} & \textbf{SKCM}
			& \\
			\midrule
			
			\multirow{2}{*}{
				\begin{tabular}[c]{@{}c@{}}
					\textbf{PCC}\\[-0.15em]
					$\uparrow$
				\end{tabular}
			}
			& Learnable state
			& $0.317$ & $\mathbf{0.367}$ & $0.129$ & $0.590$ & $0.618$
			& $0.311$ & $0.506$ & $0.414$ & $0.238$ & $0.701$
			& $0.419$ \\
			
			& \cellcolor{hestbg}\textbf{Learnable state + perturb}
			& \cellcolor{hestbg}$\mathbf{0.335}$ 
			& \cellcolor{hestbg}$0.363$ 
			& \cellcolor{hestbg}$\mathbf{0.132}$ 
			& \cellcolor{hestbg}$\mathbf{0.595}$ 
			& \cellcolor{hestbg}$\mathbf{0.621}$
			& \cellcolor{hestbg}$\mathbf{0.315}$ 
			& \cellcolor{hestbg}$\mathbf{0.516}$ 
			& \cellcolor{hestbg}$\mathbf{0.422}$ 
			& \cellcolor{hestbg}$\mathbf{0.245}$ 
			& \cellcolor{hestbg}$\mathbf{0.703}$
			& \cellcolor{hestbg}$\mathbf{0.425}$ \\
			
			\midrule
			
			\multirow{2}{*}{
				\begin{tabular}[c]{@{}c@{}}
					\textbf{MSE}\\[-0.15em]
					$\downarrow$
				\end{tabular}
			}
			& Learnable state
			& $0.517$ & $1.864$ & $\mathbf{0.410}$ & $2.671$ & $2.022$
			& $\mathbf{0.514}$ & $1.349$ & $\mathbf{0.537}$ & $\mathbf{0.296}$ & $1.993$
			& $1.217$ \\
			
			& \cellcolor{hestbg}\textbf{Learnable state + perturb}
			& \cellcolor{hestbg}$\mathbf{0.508}$ 
            & \cellcolor{hestbg}$\mathbf{1.853}$ 
			& \cellcolor{hestbg}$0.415$ 
			& \cellcolor{hestbg}$\mathbf{2.656}$ 
			& \cellcolor{hestbg}$\mathbf{2.018}$
			& \cellcolor{hestbg}$0.528$ 
			& \cellcolor{hestbg}$\mathbf{1.346}$ 
			& \cellcolor{hestbg}$0.550$ 
			& \cellcolor{hestbg}$0.305$ 
			& \cellcolor{hestbg}$\mathbf{1.872}$
			& \cellcolor{hestbg}$\mathbf{1.205}$ \\
			
			\bottomrule
		\end{tabular}
	}
\end{table*}

\subsection{Training-time Perturbation of Initial State}
\label{app:initialization}
Table~\ref{tab:initial-prior-ablation} studies the effect of training-time perturbation on the coarsest-scale initial expression state.
Since Carta is formulated as deterministic field reconstruction at evaluation, both variants use the same deterministic initialization at test time.
The only difference is whether a small stochastic perturbation is added to the learnable initial state during training.

Adding this perturbation increases the average PCC from $0.419$ to $0.425$ and decreases the average MSE from $1.217$ to $1.205$.
This suggests that training-time perturbation provides useful regularization for the coarse-to-fine reconstruction path and prevents the model from relying too strongly on a fixed initial state.
The organ-wise results show that the perturbation improves most tasks, while the deterministic learnable state remains competitive on a few organs.
Since both variants use deterministic initialization at evaluation, this comparison isolates the effect of training-time perturbation.
In Carta, we therefore use the perturbed learnable state during training and deterministic mean initialization during evaluation.

\begin{table*}[htbp]
    \centering
    \tiny
    \caption{
        \textbf{Scale-wise MSE Supervision Analysis on HEST.}
        We compare MSE supervision applied only at the finest scale
        with a variant that applies MSE supervision at every
        reconstruction scale. Best results are highlighted in bold,
        and the finest-scale-only setting is shaded.
    }
    \label{tab:scale-mse-ablation}

    \setlength{\tabcolsep}{3.9pt}
    \renewcommand{\arraystretch}{1.14}

    \resizebox{\textwidth}{!}{
        \begin{tabular}{@{}ll*{11}{c}@{}}
            \toprule
            \multirow{2}{*}{\textbf{Metric}}
            & \multirow{2}{*}{\textbf{Supervision}}
            & \multicolumn{10}{c}{\textbf{Organ}}
            & \multirow{2}{*}{\textbf{Avg.}} \\
            \cmidrule(lr){3-12}
            &
            & \textbf{CCRCC} & \textbf{COAD} & \textbf{HCC} & \textbf{IDC} & \textbf{LUNG}
            & \textbf{LYMPH} & \textbf{PAAD} & \textbf{PRAD} & \textbf{READ} & \textbf{SKCM}
            & \\
            \midrule

            \multirow{2}{*}{
                \begin{tabular}[c]{@{}c@{}}
                    \textbf{PCC}\\[-0.15em]
                    $\uparrow$
                \end{tabular}
            }
            & MSE at every scale
            & $0.296$ & $0.355$ & $0.122$ & $0.570$ & $0.617$
            & $0.305$ & $0.501$ & $0.417$ & $0.174$ & $0.694$
            & $0.405$ \\

            & \cellcolor{hestbg}\textbf{Finest-scale only}
            & \cellcolor{hestbg}$\mathbf{0.335}$
            & \cellcolor{hestbg}$\mathbf{0.363}$
            & \cellcolor{hestbg}$\mathbf{0.132}$
            & \cellcolor{hestbg}$\mathbf{0.595}$
            & \cellcolor{hestbg}$\mathbf{0.621}$
            & \cellcolor{hestbg}$\mathbf{0.315}$
            & \cellcolor{hestbg}$\mathbf{0.516}$
            & \cellcolor{hestbg}$\mathbf{0.422}$
            & \cellcolor{hestbg}$\mathbf{0.245}$
            & \cellcolor{hestbg}$\mathbf{0.703}$
            & \cellcolor{hestbg}$\mathbf{0.425}$ \\

            \midrule

            \multirow{2}{*}{
                \begin{tabular}[c]{@{}c@{}}
                    \textbf{MSE}\\[-0.15em]
                    $\downarrow$
                \end{tabular}
            }
            & MSE at every scale
            & $0.525$
            & $\mathbf{1.821}$
            & $\mathbf{0.413}$
            & $3.001$
            & $2.023$
            & $0.529$
            & $\mathbf{1.334}$
            & $\mathbf{0.532}$
            & $0.317$
            & $2.004$
            & $1.250$ \\

            & \cellcolor{hestbg}\textbf{Finest-scale only}
            & \cellcolor{hestbg}$\mathbf{0.508}$
            & \cellcolor{hestbg}$1.853$
            & \cellcolor{hestbg}$0.415$
            & \cellcolor{hestbg}$\mathbf{2.656}$
            & \cellcolor{hestbg}$\mathbf{2.018}$
            & \cellcolor{hestbg}$\mathbf{0.528}$
            & \cellcolor{hestbg}$1.346$
            & \cellcolor{hestbg}$0.550$
            & \cellcolor{hestbg}$\mathbf{0.305}$
            & \cellcolor{hestbg}$\mathbf{1.872}$
            & \cellcolor{hestbg}$\mathbf{1.205}$ \\

            \bottomrule
        \end{tabular}
    }
\end{table*}

\subsection{Scale-wise Supervision Analysis}
\label{app:scale-mse-ablation}
Intermediate-scale expression targets are obtained by aggregating spot-level measurements within spatial bins, and thus provide approximate regional summaries rather than independently measured expression profiles. Imposing direct MSE supervision on these targets may overconstrain intermediate predictions and encourage over-smoothing, potentially suppressing localized expression variations. We therefore supervise the final predictions at the original measured coordinates, allowing intermediate expression states to guide subsequent refinement without being forced to fit potentially imprecise regional averages. To validate this design choice, we compare the default objective with an alternative that applies MSE supervision at every reconstruction scale. As shown in \cref{tab:scale-mse-ablation}, the latter consistently reduces PCC across all 10 organs, lowering the average from 0.425 to 0.405, while increasing the average MSE from 1.205 to 1.250. These results support our choice to avoid direct MSE supervision at intermediate scales, as additional constraints on aggregation-derived targets do not improve final field reconstruction.

\subsection{BCT Hyperparameter Sensitivity}
\label{app:bct-sensitivity}
\begin{wraptable}{r}{0.50\textwidth}
\vspace{-8pt}
\centering
\caption{
\textbf{Sensitivity analysis of BCT hyperparameters.}
The default setting is highlighted.
}
\label{tab:bct-sensitivity}
\tiny
\setlength{\tabcolsep}{3.2pt}
\renewcommand{\arraystretch}{1.05}
\captionsetup[subtable]{font=tiny,labelfont=normalfont,skip=2pt}

\begin{subtable}[t]{0.48\linewidth}
\centering
\begin{tabular}{lcc}
\toprule
$\lambda_{\mathrm{BCT}}$ & PCC$\uparrow$ & MSE$\downarrow$ \\
\midrule
$0$     & $0.413$ & $1.225$ \\
\rowcolor{bctbluebg}
$0.001$ & $\mathbf{0.425}$ & $\mathbf{1.205}$ \\
$0.01$  & $0.417$ & $1.220$ \\
\bottomrule
\end{tabular}
\caption{$\lambda_{\mathrm{BCT}}$}
\end{subtable}
\hfill
\begin{subtable}[t]{0.48\linewidth}
\centering
\begin{tabular}{lcc}
\toprule
$\tau$ & PCC$\uparrow$ & MSE$\downarrow$ \\
\midrule
$0.5$ & $0.419$ & $1.217$ \\
\rowcolor{bctorangebg}
$1$   & $\mathbf{0.425}$ & $\mathbf{1.205}$ \\
$2$   & $0.421$ & $1.215$ \\
\bottomrule
\end{tabular}
\caption{Temperature $\tau$}
\end{subtable}

\vspace{3pt}

\begin{subtable}[t]{0.48\linewidth}
\centering
\begin{tabular}{lcc}
\toprule
$\alpha$ & PCC$\uparrow$ & MSE$\downarrow$ \\
\midrule
$0$   & $0.420$ & $1.213$ \\
\rowcolor{bctgreenbg}
$0.5$ & $\mathbf{0.425}$ & $\mathbf{1.205}$ \\
$1$   & $0.421$ & $1.211$ \\
\bottomrule
\end{tabular}
\caption{Balance $\alpha$}
\end{subtable}
\hfill
\begin{subtable}[t]{0.48\linewidth}
\centering
\begin{tabular}{lcc}
\toprule
$\beta$ & PCC$\uparrow$ & MSE$\downarrow$ \\
\midrule
$0$    & $0.422$ & $1.212$ \\
\rowcolor{bctpurplebg}
$0.05$ & $\mathbf{0.425}$ & $\mathbf{1.205}$ \\
$0.1$  & $0.423$ & $1.210$ \\
\bottomrule
\end{tabular}
\caption{Momentum $\beta$}
\end{subtable}
\vspace{-8pt}
\end{wraptable}

Table~\ref{tab:bct-sensitivity} studies the sensitivity of BCT to its key hyperparameters on HEST.
Overall, Carta remains stable under different BCT configurations, while the default setting consistently achieves the best average PCC and MSE.
When $\lambda_{\mathrm{BCT}}=0$, the model degenerates to the variant without BCT, leading to a clear drop in PCC from $0.425$ to $0.413$ and an increase in MSE from $1.205$ to $1.225$.
This confirms the effectiveness of BCT beyond point-wise regression.
Increasing $\lambda_{\mathrm{BCT}}$ to $0.01$ also degrades performance, suggesting that an overly strong BCT constraint may interfere with coordinate-level expression fitting.

For the temperature $\tau$, the best performance is obtained at $\tau=1$.
A smaller temperature makes the anchor assignment sharper, while a larger temperature produces smoother assignments; both variants slightly reduce performance.
For the balance coefficient $\alpha$, using both transport directions with $\alpha=0.5$ outperforms using either direction alone, indicating that bidirectional conditional transport is beneficial for aligning node representations and gene anchors.
Finally, the momentum coefficient $\beta=0.05$ gives the best results among the tested values.
The setting $\beta=0$ disables the momentum update of the gene-anchor prior; since the prior is initialized uniformly, this setting effectively reduces BCT to the variant without a dynamic prior.
Compared with $\beta=0$ or $\beta=0.1$, the moderate update rate $\beta=0.05$ provides a more stable dynamic prior for gene-anchor assignment.

\begin{table*}[htbp]
	\centering
	\tiny
	\caption{
		\textbf{Organ-wise ablation study of gene-anchor design in BCT on HEST.}
		We report PCC and MSE for each organ using UNI as the default pathology foundation model.
		All variants use the same training configuration and BCT hyperparameters, except for the source of gene anchors.
		Best and second-best results are highlighted in bold and underlined, respectively.
	}
	\label{tab:bct-anchor-ablation}
	
	\setlength{\tabcolsep}{3.9pt}
	\renewcommand{\arraystretch}{1.14}
	
	\resizebox{\textwidth}{!}{
		\begin{tabular}{@{}ll*{11}{c}@{}}
			\toprule
			\multirow{2}{*}{\textbf{Metric}} 
			& \multirow{2}{*}{\textbf{BCT variant}}
			& \multicolumn{10}{c}{\textbf{Organ}} 
			& \multirow{2}{*}{\textbf{Avg.}} \\
			\cmidrule(lr){3-12}
			& 
			& \textbf{CCRCC} & \textbf{COAD} & \textbf{HCC} & \textbf{IDC} & \textbf{LUNG}
			& \textbf{LYMPH} & \textbf{PAAD} & \textbf{PRAD} & \textbf{READ} & \textbf{SKCM}
			& \\
			\midrule
			
			\multirow{3}{*}{
				\begin{tabular}[c]{@{}c@{}}
					\textbf{PCC}\\[-0.15em]
					$\uparrow$
				\end{tabular}
			}
			& w/o BCT
			& $0.316$ & $0.352$ & $\underline{0.119}$ & $0.576$ & $0.612$
			& $0.308$ & $\mathbf{0.517}$ & $0.412$ & $0.238$ & $0.683$
			& $0.413$ \\
			
			& Separate anchors
			& $\underline{0.330}$ & $\mathbf{0.366}$ & $0.118$ & $\underline{0.585}$ & $\underline{0.618}$
			& $\underline{0.310}$ & $0.505$ & $\underline{0.421}$ & $\mathbf{0.248}$ & $\underline{0.701}$
			& $\underline{0.420}$ \\
			
			& \cellcolor{hestbg}\textbf{Prediction-head anchors}
			& \cellcolor{hestbg}$\mathbf{0.335}$
			& \cellcolor{hestbg}$\underline{0.363}$
			& \cellcolor{hestbg}$\mathbf{0.132}$
			& \cellcolor{hestbg}$\mathbf{0.595}$
			& \cellcolor{hestbg}$\mathbf{0.621}$
			& \cellcolor{hestbg}$\mathbf{0.315}$
			& \cellcolor{hestbg}$\underline{0.516}$
			& \cellcolor{hestbg}$\mathbf{0.422}$
			& \cellcolor{hestbg}$\underline{0.245}$
			& \cellcolor{hestbg}$\mathbf{0.703}$
			& \cellcolor{hestbg}$\mathbf{0.425}$ \\
			
			\midrule
			
			\multirow{3}{*}{
				\begin{tabular}[c]{@{}c@{}}
					\textbf{MSE}\\[-0.15em]
					$\downarrow$
				\end{tabular}
			}
			& w/o BCT
			& $\mathbf{0.498}$ & $\mathbf{1.822}$ & $0.420$ & $2.760$ & $\mathbf{1.987}$
			& $0.556$ & $\mathbf{1.294}$ & $\underline{0.566}$ & $\underline{0.304}$ & $2.040$
			& $1.225$ \\
			
			& Separate anchors
			& $\underline{0.499}$ & $\underline{1.837}$ & $\mathbf{0.403}$ & $\underline{2.666}$ & $2.040$
			& $\mathbf{0.517}$ & $\underline{1.345}$ & $0.568$ & $\mathbf{0.284}$ & $\underline{1.957}$
			& $\underline{1.212}$ \\
			
			& \cellcolor{hestbg}\textbf{Prediction-head anchors}
			& \cellcolor{hestbg}$0.508$
			& \cellcolor{hestbg}$1.853$
			& \cellcolor{hestbg}$\underline{0.415}$
			& \cellcolor{hestbg}$\mathbf{2.656}$
			& \cellcolor{hestbg}$\underline{2.018}$
			& \cellcolor{hestbg}$\underline{0.528}$
			& \cellcolor{hestbg}$1.346$
			& \cellcolor{hestbg}$\mathbf{0.550}$
			& \cellcolor{hestbg}$0.305$
			& \cellcolor{hestbg}$\mathbf{1.872}$
			& \cellcolor{hestbg}$\mathbf{1.205}$ \\
			
			\bottomrule
		\end{tabular}
	}
\end{table*}

\subsection{Gene-anchor Design in BCT}
\label{app:bct-anchor-ablation}
Table~\ref{tab:bct-anchor-ablation} presents an organ-wise comparison of different gene-anchor designs in BCT. 
Compared with the variant without BCT, using separate learnable anchors improves
the average PCC from $0.413$ to $0.420$ and reduces the average MSE from $1.225$ to
$1.212$, indicating that an explicit gene-anchor space provides useful structure
beyond point-wise regression. 
However, the default design that uses the prediction-head weights as gene anchors achieves the best overall performance, further improving the average PCC to $0.425$ and reducing the average MSE to $1.205$.
This suggests that the latent directions directly responsible for gene-wise expression prediction provide a more consistent anchor space than separately learned anchors.

The organ-wise results also show that lower MSE does not always correspond to higher PCC.
For example, in CCRCC, COAD and LUNG, the variant without BCT obtains competitive or even lower MSE, but its PCC is clearly lower than that of the BCT-based variants.
This indicates that point-wise MSE alone cannot fully supervise the reconstruction of spatial expression patterns.
While MSE mainly measures absolute prediction error, PCC better reflects whether the predicted expression preserves the relative spatial variation across spots.
Therefore, the improvement brought by BCT is not merely a reduction of point-wise error, but a better preservation of expression structures and gene-wise variation patterns.

\begin{figure}[htbp]
    \centering
    \includegraphics[width=0.74\linewidth]
    {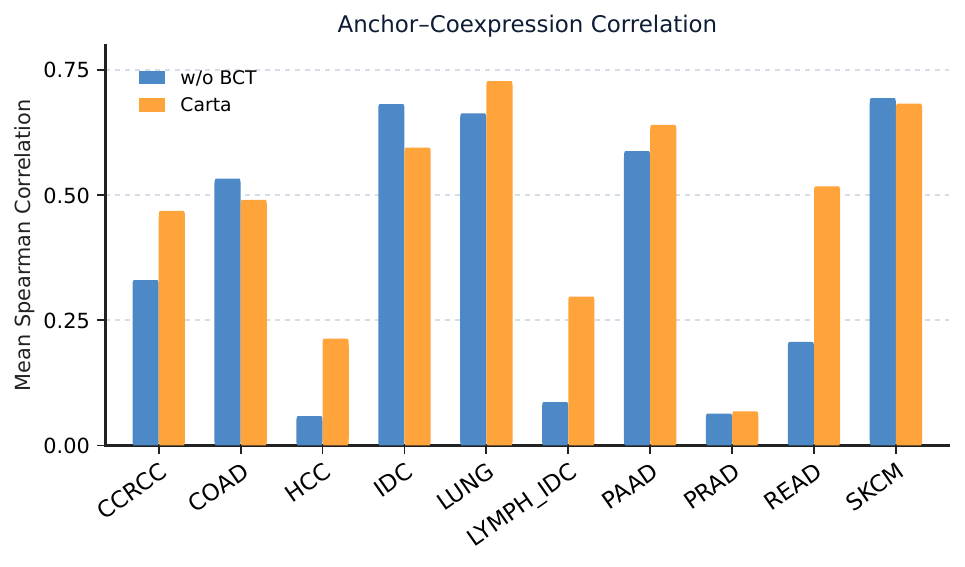}
    \caption{
\textbf{Association between gene-anchor geometry and gene co-expression across cancer types.}
For each patient-disjoint HEST fold, we calculate the Spearman correlation between pairwise cosine similarities of the learned gene anchors and gene co-expression computed exclusively from the corresponding training spots.
The bars report the mean correlation across folds belonging to the same cancer type for Carta and its variant without BCT.
}
    \label{fig:anchor-coexpression}
\end{figure}

\subsection{Relation to Gene Co-expression}
\label{app:anchor-coexpression}
As shown in Figure~\ref{fig:anchor-coexpression}, the learned gene-anchor geometry exhibits positive mean correlations with gene co-expression across all evaluated cancer types.
The association is particularly strong for LUNG, SKCM, and PAAD, while relatively weaker correlations are observed for PRAD, HCC, and LYMPH\_IDC.
Compared with the variant without BCT, Carta achieves higher mean correlations in seven of the ten cancer types, with particularly clear improvements on CCRCC, HCC, LYMPH\_IDC, and READ.
Overall, the mean correlation increases from 0.39 without BCT to 0.47 with BCT.
These results suggest that the expression-supervised gene-anchor space used by BCT reflects gene co-expression structure, despite substantial variation in correlation strength across cancer types.
This observation is consistent with the gene-aware design of BCT, where each gene is represented by a dedicated anchor and the bidirectional assignment encourages each gene anchor to receive support from spatial nodes rather than concentrating the matching on a small set of frequently activated gene responses.

\begin{table*}[htbp]
	\centering
	\caption{
		\textbf{Efficiency comparison on HEST.}
		We report trainable parameters, inference time normalized to 10,000 spots, and average PCC for representative neural methods.
		Feature extraction time is excluded when applicable.
	}
	\label{tab:efficiency-comparison}
	\small
	\setlength{\tabcolsep}{7.5pt}
	\renewcommand{\arraystretch}{1.15}
	
	\resizebox{\textwidth}{!}{
		\begin{tabular}{@{}llccc@{}}
			\toprule
			\multirow{2}{*}{\textbf{Method}}
			& \multirow{2}{*}{\textbf{Prediction paradigm}}
			& \multicolumn{2}{c}{\textbf{Computational cost}}
			& \textbf{Performance} \\
			\cmidrule(lr){3-4}
			\cmidrule(l){5-5}
			& 
			& \textbf{Trainable Params. $\downarrow$}
			& \textbf{Time / 10K spots $\downarrow$}
			& \textbf{Avg. PCC $\uparrow$} \\
			\midrule
			
			STNet
			& Spot-wise neural regression
			& $\mathbf{0.102}$M
			& $0.121$s
			& $0.317$ \\
			
			HisToGene
			& Slide-level Transformer regression
			& $169.075$M
			& $0.397$s
			& $0.213$ \\
			
			BLEEP
			& Contrastive retrieval-based prediction
			& $0.672$M
			& $\mathbf{0.036}$s
			& $0.350$ \\
			
			Stem
			& Diffusion-based generation
			& $7.474$M
			& $318.311$s
			& $0.359$ \\
			
			STFlow
			& Flow matching generation
			& $1.148$M
			& $0.508$s
			& $0.401$ \\
			
			\midrule[-1pt]
			\cellcolor{hestbg}\textbf{Carta}
			& \cellcolor{hestbg}\textbf{Scale-wise expression-field reconstruction}
			& \cellcolor{hestbg}$4.526$M
			& \cellcolor{hestbg}$0.230$s
			& \cellcolor{hestbg}$\mathbf{0.425}$ \\
			
			\bottomrule
		\end{tabular}
	}
\end{table*}

\subsection{Efficiency Analysis}
\label{app:efficiency}
Before analyzing the results, we clarify the evaluation protocol for efficiency comparison.
Trainable parameters only include modules updated during training under each method's actual training configuration.
For runtime measurement, we exclude image or pathology feature extraction when applicable and report the inference time normalized to 10,000 spots.
For generative baselines, Stem is evaluated with 1000 DDPM denoising steps and one generated sample per spot, while STFlow uses 5 flow-matching refinement steps.
Thus, the reported runtime mainly reflects the cost of the prediction module rather than the frozen histology encoder.

Table~\ref{tab:efficiency-comparison} compares the computational cost and predictive accuracy of representative neural methods.
Lightweight methods such as STNet and BLEEP have small trainable modules and fast inference, but their average PCC remains limited, suggesting that simple spot-wise regression or retrieval-based prediction is insufficient for capturing spatial expression patterns.
Stem improves the prediction accuracy by introducing diffusion-based generation, but its 1000-step DDPM sampling leads to a much higher inference cost.
STFlow largely alleviates this issue with 5-step flow-matching refinement, reducing the runtime from $318.311$s to $0.508$s while further improving the average PCC.
Carta takes another step by reconstructing the expression field in a scale-wise manner.
Compared with STFlow, Carta increases the average PCC from $0.401$ to $0.425$ and reduces the normalized inference time from $0.508$s to $0.230$s.
This indicates that Carta does not simply trade efficiency for accuracy, but achieves stronger prediction with a moderate computational budget.

\section{Qualitative and Further Analysis}
\label{app:qualitative-analysis}

This section presents an expanded collection of qualitative visualizations of Carta. We extend the comparisons in the main text to a broader set of biomarker genes, providing additional expression maps and comparisons with baseline methods. Across these examples, we further illustrate Carta's coarse-to-fine reconstruction process at intermediate scales, showing how regional expression estimates are progressively refined into predictions at native spatial resolution. Together, these visualizations provide a more comprehensive qualitative assessment of Carta's predictive performance and multi-scale expression field reconstruction.

\subsection{Additional Biomarker Expression Maps}
\label{app:more_vis_maps}
In this section, we provide additional qualitative comparisons on four representative biomarkers, including KRTDAP, PTPRC, MET, and UBE2C.
Compared with Figures~\ref{fig:concept} and~\ref{fig:biomarker} in the main text, we include more baseline methods and additional biomarker genes for a more comprehensive supplementary comparison.
These visualizations complement the quantitative results in the main text by showing how different methods reconstruct spatial expression patterns at the tissue-section level. 

\begin{figure}[p!]
    \centering
    \includegraphics[width=0.75\linewidth]{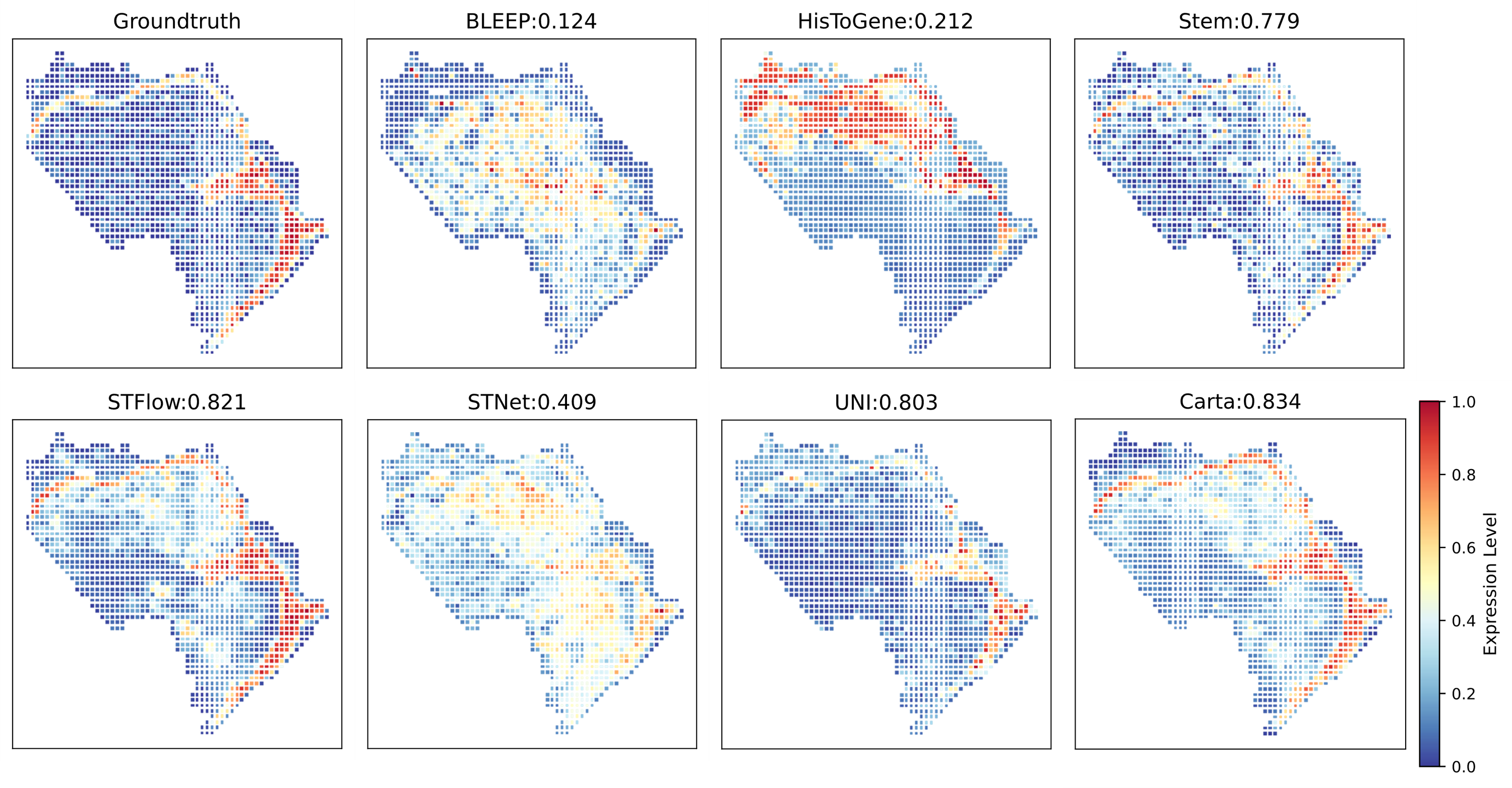}
    \caption{Qualitative comparison on KRTDAP expression prediction at TENX115.}
    \label{fig:app_vis_KRTDAP}
\end{figure}

\begin{figure}[p!]
    \centering
    \includegraphics[width=0.75\linewidth]{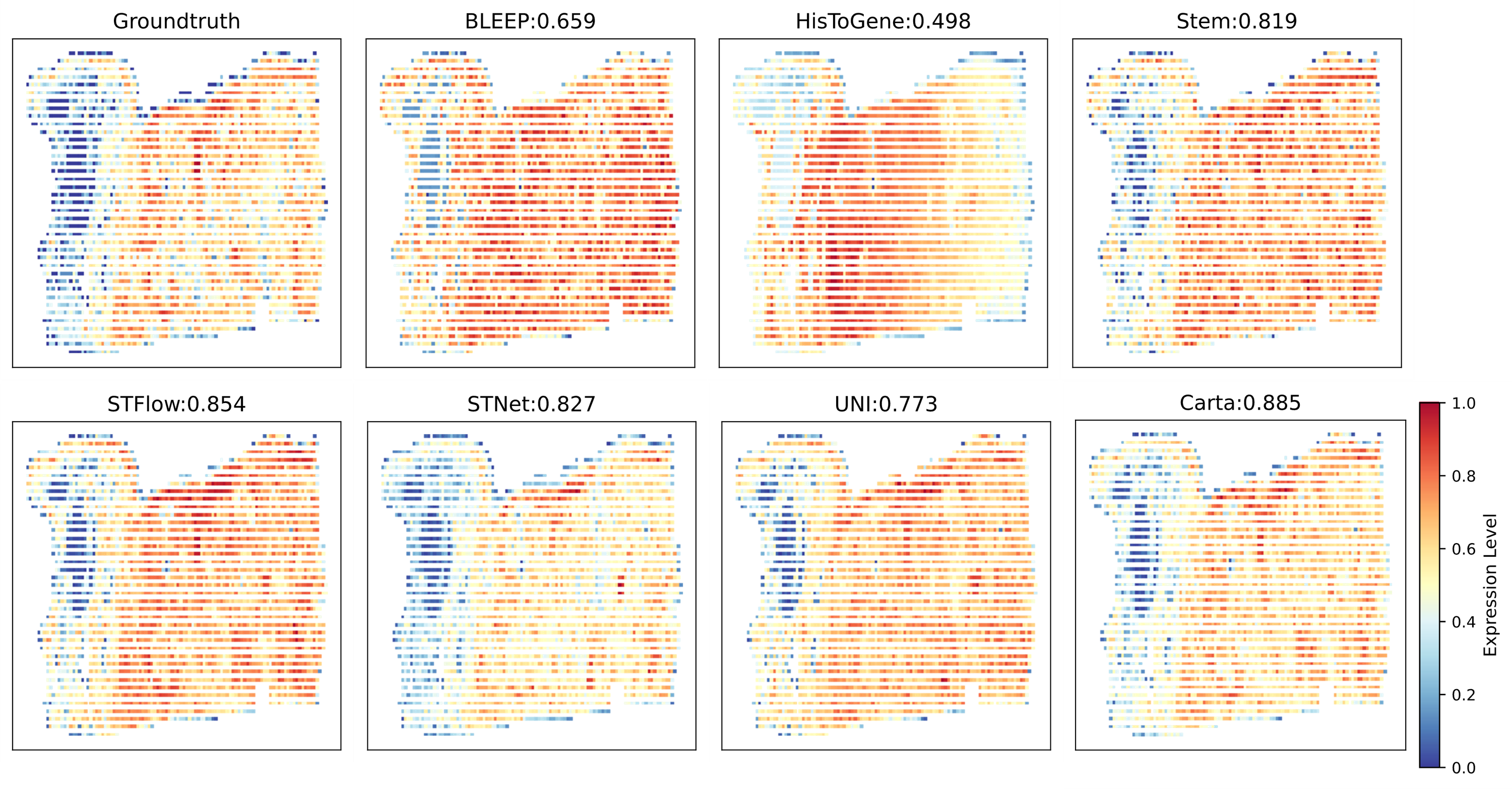}
    \caption{Qualitative comparison on PTPRC expression prediction at TENX141.}
    \label{fig:app_vis_PTPRC}
\end{figure}

\begin{figure}[p!]
    \centering
    \includegraphics[width=0.75\linewidth]{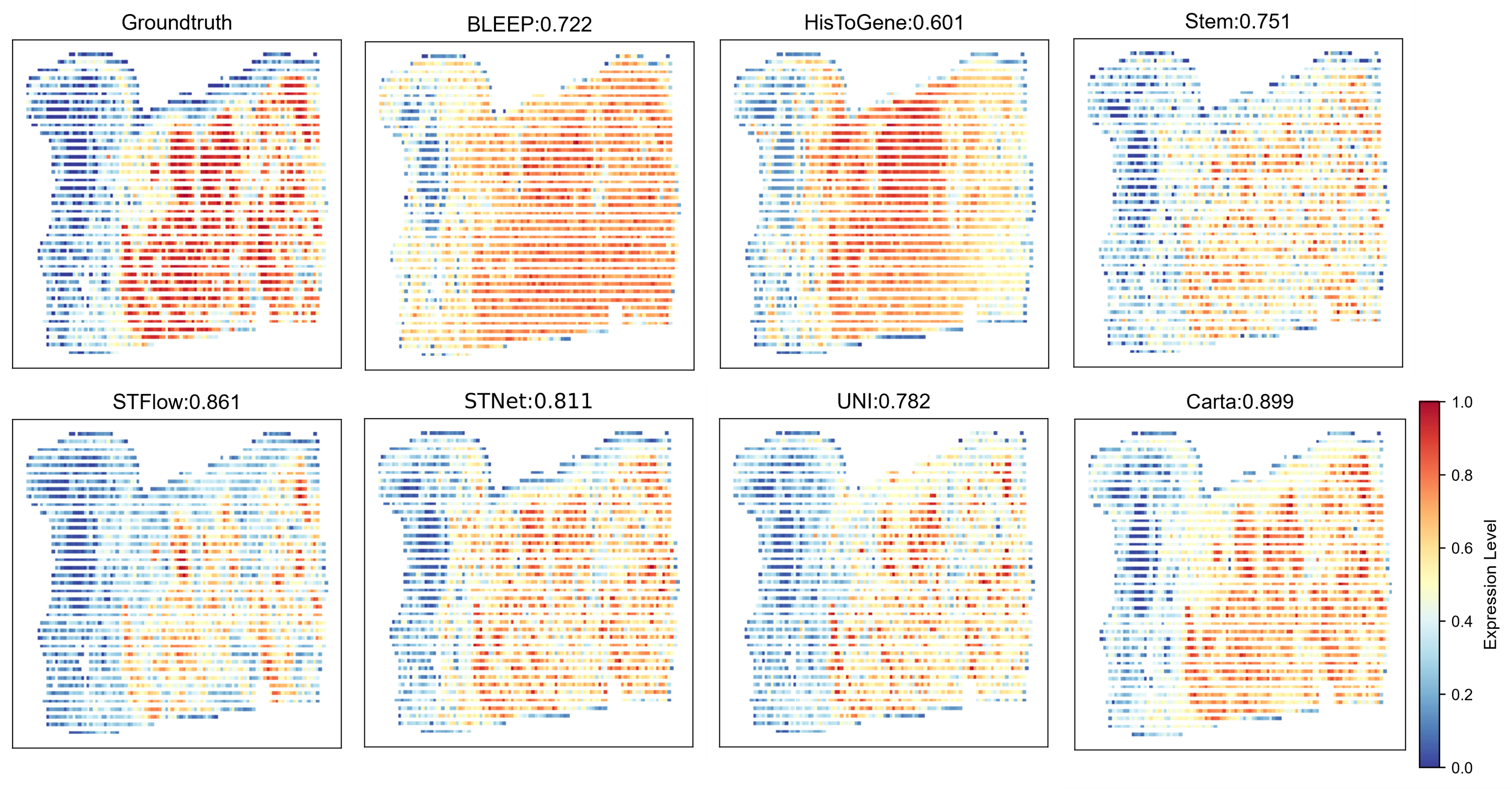}
    \caption{Qualitative comparison on MET expression prediction at TENX141.}
    \label{fig:app_vis_MET}
\end{figure}

\begin{figure}[p!]
    \centering
    \includegraphics[width=0.75\linewidth]{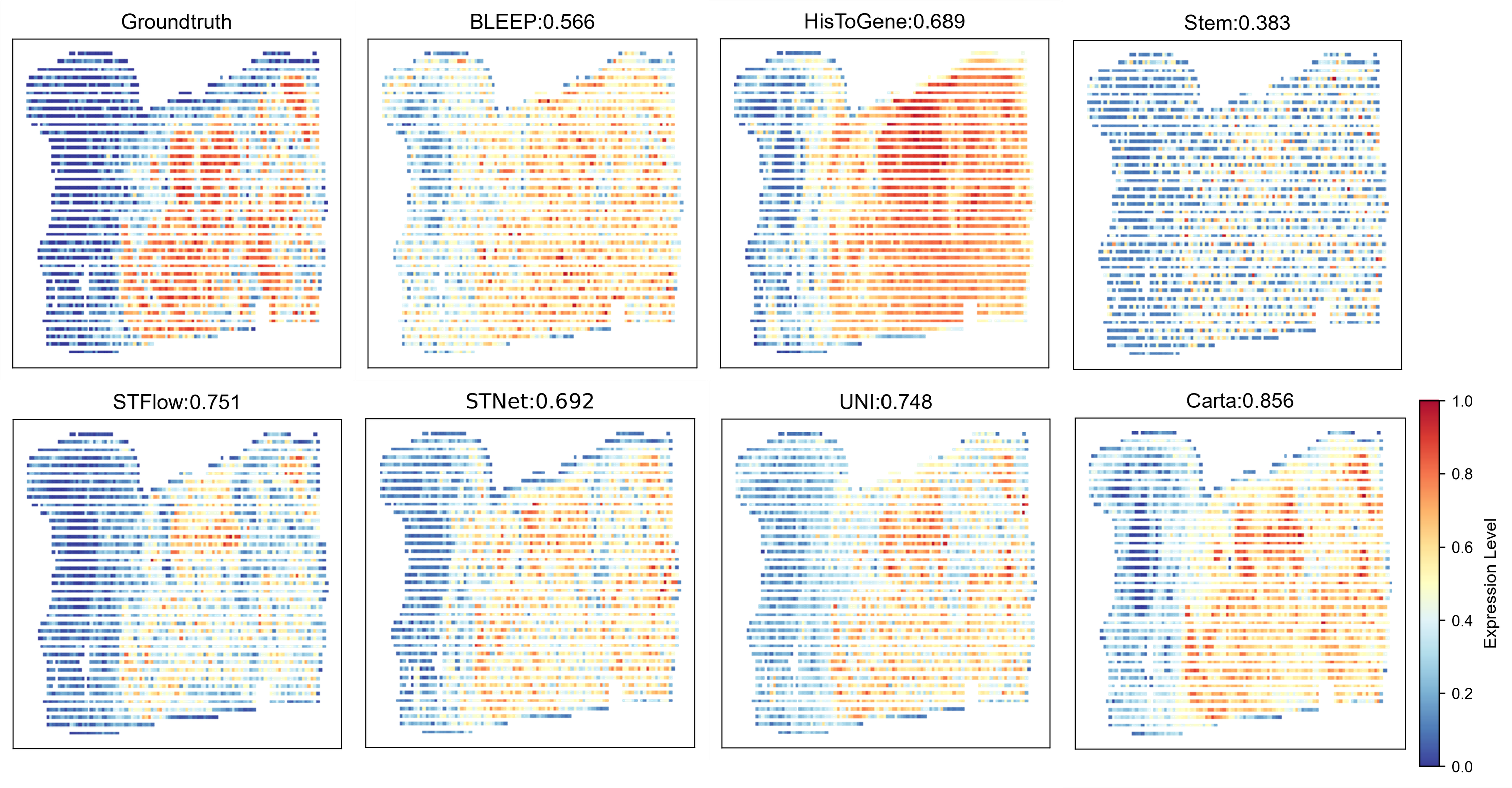}
    \caption{Qualitative comparison on UBE2C expression prediction at TENX141.}
    \label{fig:app_vis_UBE2C}
\end{figure}

As shown in Figures~\ref{fig:app_vis_KRTDAP}--\ref{fig:app_vis_UBE2C}, several baselines recover part of the global expression trend, especially for biomarkers with broad activation regions, but their predictions are less consistent across genes.
For the sparse KRTDAP pattern, some methods miss boundary-localized activation or introduce signals in unrelated tissue areas.
For PTPRC, MET, and UBE2C, the shared tissue outline but different ground-truth expression fields make it possible to examine whether a model simply follows tissue morphology or truly adapts to gene-specific spatial patterns.
In these cases, some baselines produce overly broad activation maps or smooth out the contrast between high- and low-expression regions.

Carta gives more stable predictions across these settings.
It better preserves the boundary-associated activation for KRTDAP and adapts to distinct biomarker-specific patterns for PTPRC, MET, and UBE2C under the same morphology.
Although some baselines also produce reasonable maps in certain cases, Carta more consistently matches the measured expression fields in both spatial localization and expression contrast.
These observations are consistent with the quantitative biomarker results and further support the benefit of scale-wise expression field reconstruction.

\subsection{Visualization of Coarse-to-Fine Reconstruction}
\label{app:vis-coarse-to-fine}
\begin{figure}[p!]
    \centering
    \includegraphics[width=0.75\linewidth]{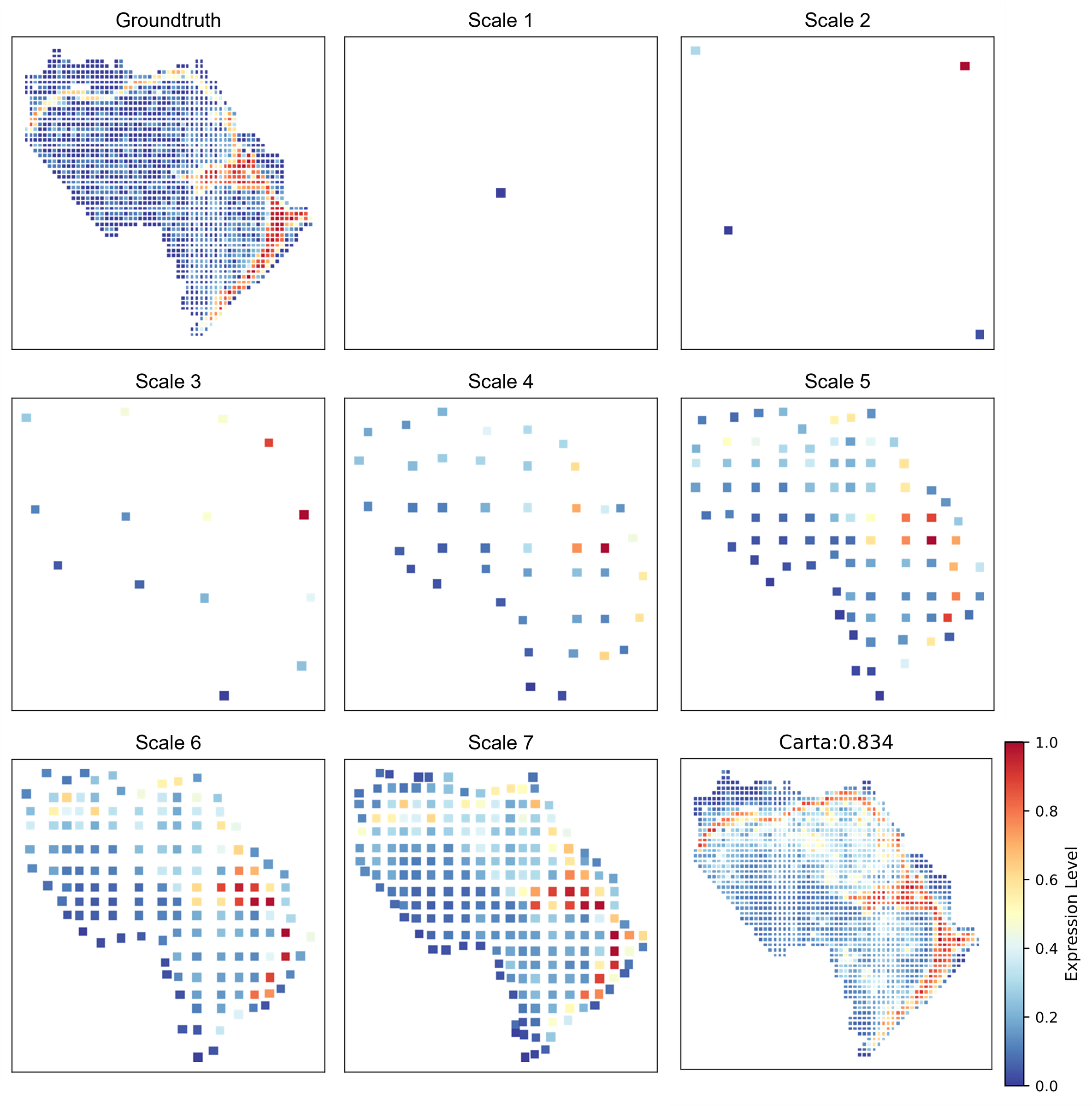}
    \caption{
    Coarse-to-fine reconstruction of KRTDAP expression maps by Carta at TENX115.
    }
    \label{fig:vis_scale_KRTDAP}
\end{figure}

\begin{figure}[p!]
    \centering
    \includegraphics[width=0.75\linewidth]{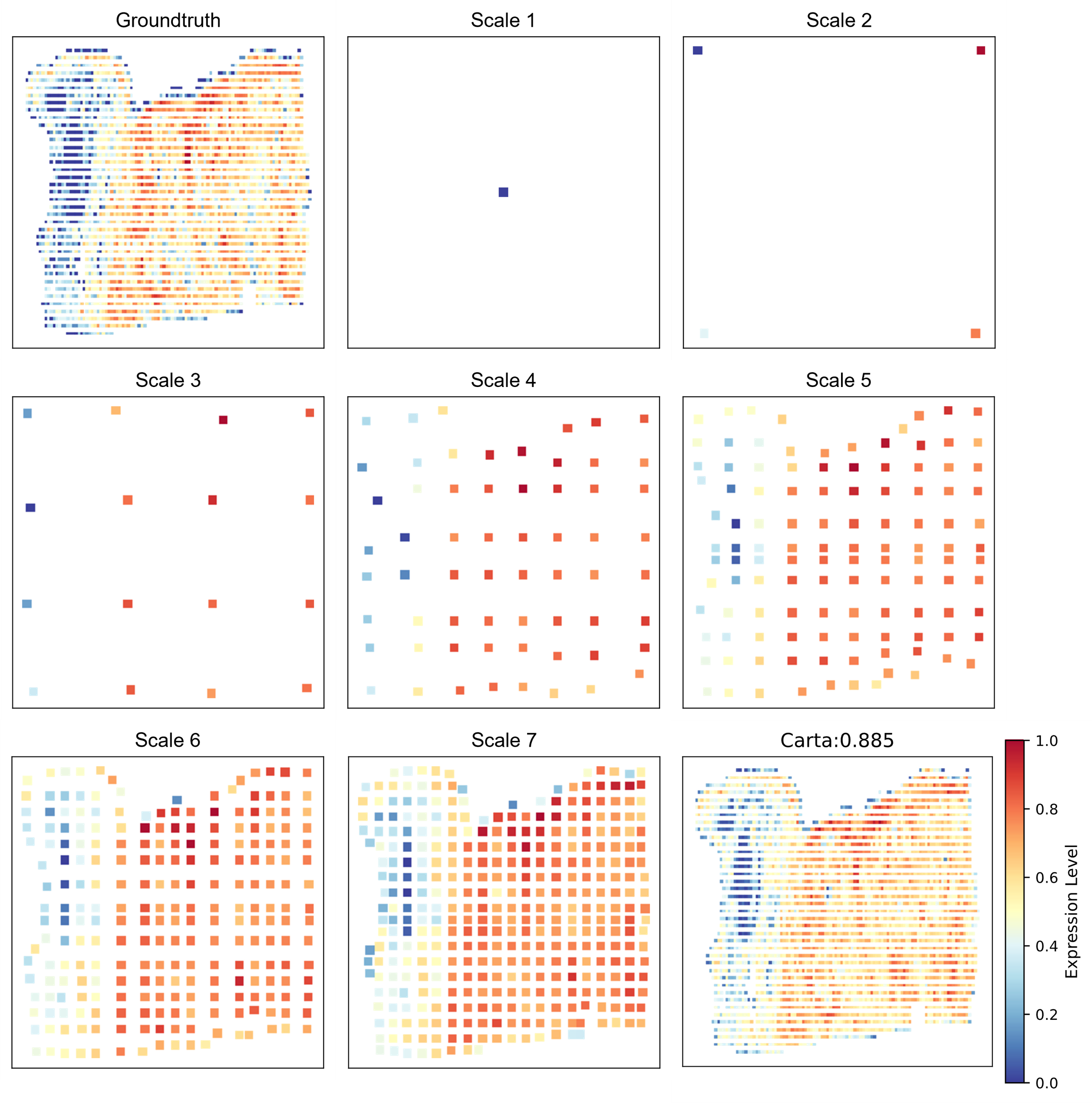}
    \caption{
    Coarse-to-fine reconstruction of PTPRC expression maps by Carta at TENX141.
    }
    \label{fig:vis_scale_PTPRC}
\end{figure}

\begin{figure}[p!]
    \centering
    \includegraphics[width=0.75\linewidth]{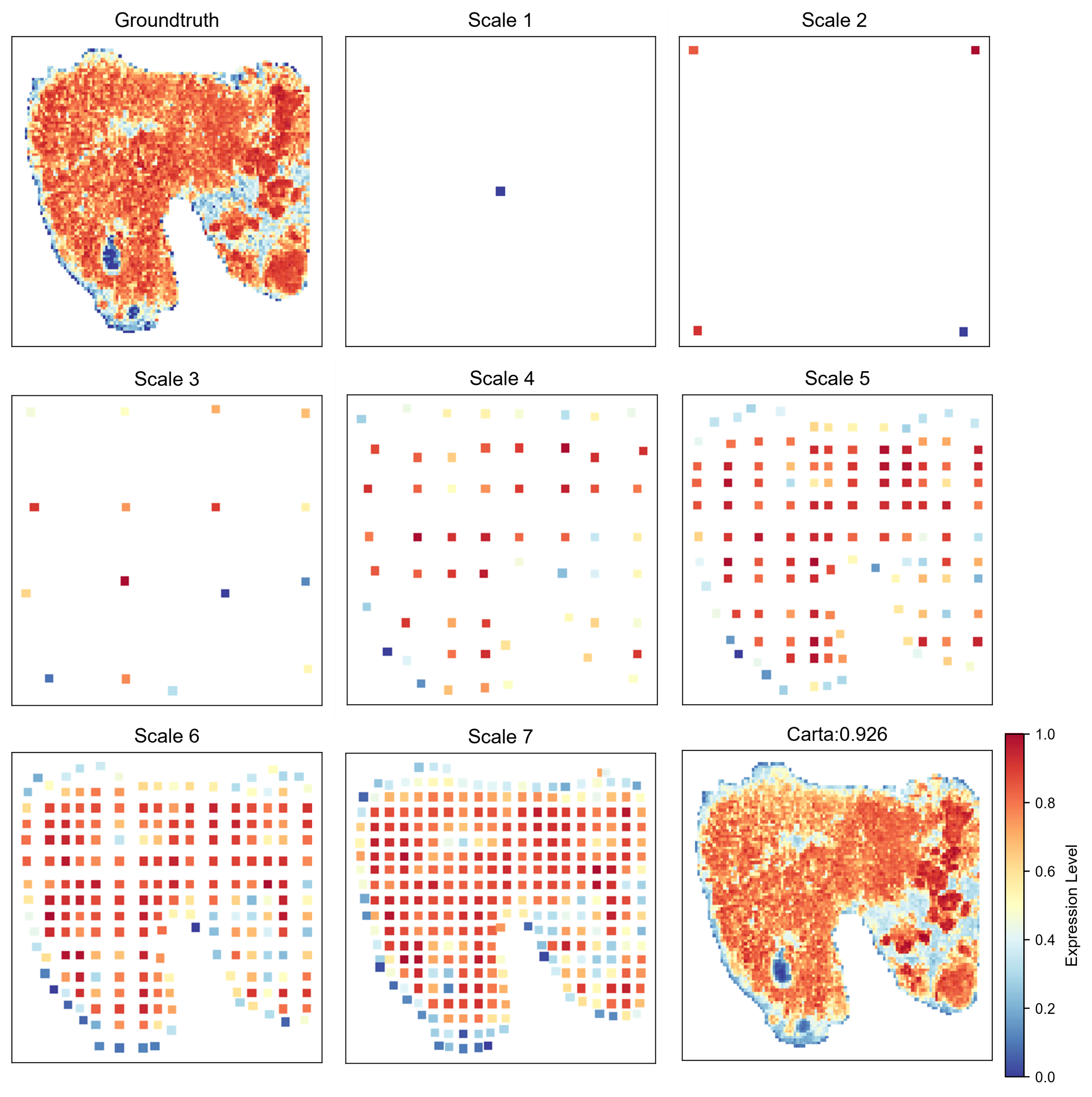}
    \caption{
    Coarse-to-fine reconstruction of ERBB2 expression maps by Carta at TENX95.
    }
    \label{fig:vis_scale_ERBB2}
\end{figure}

Figures~\ref{fig:vis_scale_KRTDAP}--\ref{fig:vis_scale_ERBB2} visualize the coarse-to-fine reconstruction process of Carta for representative biomarkers.
At the earliest scales, only a small number of regional nodes are available, so the maps mainly encode coarse expression priors rather than detailed spatial patterns.
As the scale increases, the predicted expression field gradually recovers the tissue outline and forms gene-specific activation regions.
For ERBB2 and PTPRC, which show broad expression patterns, the intermediate scales first capture the global high- and low-expression trends and then refine local variations at later stages.
For KRTDAP, whose expression is more localized, high-expression regions emerge progressively near the boundary-associated areas instead of appearing as isolated noisy spots.
The final native-resolution maps further sharpen these structures and better align with the measured expression fields.

These visualizations illustrate the role of scale-wise autoregression in Carta.
Coarse scales provide regional expression context, while finer scales refine the prediction using local histology and spatial neighborhood information.
This process differs from direct spot-wise regression, where each measured location is fitted independently and intermediate field structure is not explicitly modeled.
The observed refinement behavior supports our formulation of histology-based gene expression prediction as morphology-conditioned expression field reconstruction.

\section{Limitations and Broader Impact}
\label{app:limitations}




\paragraph{Limitations.}
Although Carta consistently improves spatial gene expression prediction across two benchmarks, several limitations remain.

First, our current evaluation focuses on task-specific top-50 highly variable genes rather than full-transcriptome prediction. Extending Carta to substantially larger gene sets would require further study of scalability and gene-level generalization.

Second, Carta uses a coordinate-derived hierarchy together with fixed interpolation between scales. While this design is effective for the assay layouts considered in our experiments, different spatial geometries, resolutions, or cross-platform settings may require adapting the hierarchy construction and interpolation scheme.

Third, Carta relies on frozen pathology foundation-model features to represent histology patches. Although these features provide strong morphology-aware representations, prediction quality can still depend on the domain coverage and representation quality of the underlying pathology encoder.

Finally, the scale-wise autoregressive reconstruction process introduces additional computation compared with direct single-scale prediction. In practice, Carta remains efficient because the hierarchy is compact and the pathology encoder is frozen, but very large spatial layouts may still increase memory and inference costs.

\paragraph{Broader impact.}
Carta may help reduce the cost of exploratory spatial molecular profiling by estimating spatial gene expression from routinely available histology images.
This could support hypothesis generation and help researchers select samples for follow-up spatial transcriptomics experiments.
However, the predictions are computational estimates rather than measured molecular profiles.
If used without biological validation, incorrect predictions may mislead downstream interpretation, especially in clinically sensitive settings.
Therefore, Carta should be used as a research tool for exploratory analysis rather than as a replacement for molecular assays or clinical decision making.

\FloatBarrier


\end{document}